%% file: ignore1corr.tex
\documentstyle[proceed,times,chicagob]{article}

\newcommand{\cP}{{\cal P}}
\newcommand{\cD}{{\cal D}}
\newcommand{\cX}{{\cal X}}
\newcommand{\cY}{{\cal Y}}
\newcommand{\cZ}{{\cal Z}}
\newcommand{\Lbayes}{\mbox{$L_{\mbox{\small Bayes}}$}}
\input{defn}

\title{When Ignorance is Bliss}
\author{{\bf Peter D. Gr{\"u}nwald}\\
CWI, P.O. Box 94079 \\ 1090 GB Amsterdam\\
pdg@cwi.nl\\
http://www.grunwald.nl
\And
{\bf Joseph Y.\ Halpern}\\
Cornell University\\
Ithaca, NY 14853\\
halpern@cs.cornell.edu\\
http://www.cs.cornell.edu/home/halpern
}
\date{}

\begin{document}
\maketitle
\begin{abstract}
It is commonly-accepted wisdom that more information is better, and that
information should never be ignored. Here we argue,
using both a Bayesian and a non-Bayesian analysis,
that in some situations
you are better off ignoring information if your uncertainty is represented
by a set of probability measures. These include situations in which the
information {\em is\/} relevant for the prediction task at hand. 
In the non-Bayesian
analysis, we show how ignoring information avoids {\em dilation}, the
phenomenon that additional pieces of information sometimes lead to an
increase in uncertainty. In the Bayesian analysis, we show that for small
sample sizes and certain prediction tasks, the Bayesian posterior based on
a noninformative prior 
yields
worse predictions than simply ignoring the given information.
\end{abstract}
\section{INTRODUCTION}
\label{sec:introduction}
It is commonly-accepted wisdom that more information is better, and that
information should never be ignored.  Indeed, this has been formalized
in a number of ways in a Bayesian framework, where uncertainty is
represented by a probability measure  \cite{Good67,RaiffaS61}.
In this paper, we argue that occasionally you are
better off ignoring information if your uncertainty is represented 
by a set of probability measures.  
Related observations have been
made by Seidenfeld \citeyear{Seidenfeld04}; we compare our work to his
in Section~\ref{sec:related}.

For definiteness, we focus on a relatively simple setting.  
Let $X$ be a random variable taking values in some set $\cX$, and let
$Y$ be a random variable taking values in some set $\cY$.  The goal of
an agent is to choose an action whose 
utility
depends only on the value of $Y$, after having observed the value of
$X$.  We further assume that, before making the observation, the agent
has a prior $\Pr_Y$ on the value of $Y$.  If the agent actually had a
prior on the joint distribution of $X$ and $Y$, then the obvious thing to
do would be to condition on the observation, to get the best estimate of
the value of $Y$.  But we are interested in situations where the agent
does not have a single prior on the joint distributions, but a  
family of priors
$\cP$.
As the following example shows, this is a situation
that arises often.

\xam Consider a doctor who is trying to decide if a patient has a 
flu
or tuberculosis.  The doctor
then learns the patient's address.  The doctor knows that the patient's
address may be correlated with disease (tuberculosis may be more
prevalent in some parts of the city than others), but does not know the
correlation (if any) at all.  In this case, the random variable $Y$ is
the disease that the patient has, $\cY = \{$flu, tuberculosis$\}$, 
and $X$ is the neighborhood in which the agent lives.  The doctor is
trying to choose a treatment.  The effect of the treatment depends only
on the value of $Y$.
Under these circumstances, many doctors would simply not take the patient's
address into account, thereby 
ignoring relevant information. In this paper we show that 
this commonly-adopted strategy is often quite sensible.
\exam

There is a relatively obvious sense in which ignoring information is the
right thing to do. 
Let $\cP$ be the set of all
joint distributions on $\cX \times \cY$ whose marginal on $Y$ is
$\Pr_Y$. $\cP$ represents the set of distributions compatible with the
agent's knowledge. Roughly speaking, 
if $a^*$ is the best action
given just the prior $\Pr_Y$, then 
then $a^*$ gives the same payoff for all joint distribution $\Pr \in 
\cP$ (since they all have marginal $\Pr_Y$)
We can show that every
other action $a'$ will do worse than $a^*$ against {\em some\/} joint
distribution $\Pr \in \cP$.
Therefore, ignoring
the information leads one to adopt the minimax optimal decision. This
idea is formalized as Proposition~\ref{pro:minimax} in
Section~\ref{sec:nonbayes}, where we also show that
ignoring information compares very favorably to the ``obvious'' way of
updating the set of measures $\cP$.
\commentout{
[[PETER, THIS MUST BE CLOSELY
RELATED TO USING MMEU: THAT IS, CHOOSING THE ACTION THAT GIVES THE BEST
WORST-CASE EXPECTATION WITH RESPECT TO ALL THE PROBABILITY ON $\cX \times
\cY$.  THIS IS WHAT SEIDENFELD IS LOOKING AT, SO WE SHOULD TRY TO MAKE
THE CONNECTION PRECISE.]]
}
Proposition~\ref{pro:minimax} makes 
three important assumptions:
\begin{enumerate}
\item There is no (second-order) distribution on the set of
  probabilities $\cP$. 
\item $\cP$ contains {\em all\/} probability distributions 
on $\cX \times \cY$ whose marginal is $\Pr_Y$.
\item The ``goodness'' of an action $a$ is measured by some loss or
  utility that---although it may be unknown to the agent at the time
  of updating---is {\em fixed}. In particular, it does not depend on the
observed 
  value of $X$.
\end{enumerate}
In the remainder of the paper, we investigate 
the effect of dropping these assumptions.
In Section~\ref{sec:bayes}, we consider what happens if we assume some
probability distribution on the set $\cP$ of probabilities.
The obvious question is which one to use.
We have to distinguish between purely subjective Bayesian approaches
and so-called ``noninformative'', ``pragmatic'', or 
``objective'' Bayesian approaches \cite{BernardoS94}, which are based
on adopting so-called
``non-informative priors''.
We show that for a large class of such priors, including
the uniform distribution and Jeffreys' prior, 
using
the Bayesian posterior may lead to worse
decisions than using 
the prior
$\Pr_Y$; that is, we may be better off
ignoring information rather than conditioning on a noninformative prior;
see Examples~\ref{ex:bayes1} and~\ref{ex:bayes3}.  
In these examples, the posterior is based on a relatively small sample.
Of course, as the sample
grows larger, then 
using any reasonable prior will result in a posterior that converges to
the true distribution.
This follows directly from standard Bayesian consistency theorems
\cite{Ghosal98}. 

In Section~\ref{sec:bookies} we investigate 
the effect of dropping the second and third assumptions.  
We show that once there is
partial information about the relationship between $X$ and $Y$ (so that
$\cP$ is a strict subset of the set of all probability distributions 
on $\cX \times \cY$ whose marginal is $\Pr_Y$), then the right thing to
do becomes sensitive to the 
kind of ``bookie'' or ``adversary'' that the agent can be viewed as
playing against (cf.~\cite{HT}). 
We consider some related work, particularly that of
Seidenfeld~\citeyear{Seidenfeld04}, in Section~\ref{sec:related}.
Our focus in this paper is on optimality in the minimax sense.
It is not clear that this is the most appropriate notion of optimality. 
Indeed, Seidenfeld explicitly argues that it is not, and 
the analysis in  Section~\ref{sec:bookies} suggests
that there are situations when
ignoring information is a reasonable thing to do, even though this is
not the minimax approach.  We discuss alternative notions on optimality
in Section~\ref{sec:related}.  We conclude with further discussion in
Section~\ref{sec:discussion}. 
\commentout{
If the adversary can completely determine the loss function,
then ignoring information can have 
catastrophic effects (as can any other approach to making
decisions). On the other hand, if the adversary's choice of loss
function is 
constrained in some natural ways, ignoring information may be
preferable to taking information into account, even if the agent knows
a lot more than merely the marginal $\Pr_Y$.

}
\section{WHEN IGNORING HELPS: A NON-BAYESIAN ANALYSIS}
\label{sec:nonbayes}
In this section, we formalize our problem in a non-Bayesian setting.
We then show that, in this setting, under some pragmatic assumptions,
ignoring information is a sensible strategy. 
We also show that ignoring information compares  
favorably to the standard approach of working with sets 
of measures on $\cX \times \cY$. 

As we said, we are interested in an agent who must choose some action
from a set $\cA$, where the 
loss 
of the action depends only on the
value of a random variable $Y$, which takes values in $\cY$.  
We assume that with each action $a \in \cA$ and value $y \in
\cY$ is associated some loss to the agent.  (The losses can be negative,
which amounts to a gain.)  Let $L: \cY \times \cA \rightarrow \IR
\cup \{ \infty \}$ be the loss function.\footnote{We could equally well use
utilities, which can be viewed as a positive measure of gain.  Losses
seem to be somewhat more standard in this literature.}
For ease of exposition, we assume in this paper that $\cA$ is finite.

For every action $a \in \cA$, let $L_a$ be the random variable on
$\cY$ such that $L_a(y) = L(y,a)$.  
Since $\cA$ is assumed to be finite, for
every distribution $\Pr_Y$ on $\cY$, there is 
a (not necessarily unique)  action $a^* \in \cA$ that achieves minimum
expected loss, 
that is,
\begin{equation}
\label{eq:bayesacta}
\inf_{a \in \cA}
(E_{\Pr_Y} [ L_a])= E_{\Pr_Y} [ L_{a^*}]
\end{equation}
If all the agent knows is $\Pr_Y$, then it seems reasonable for the
agent to choose an action $a^*$ that minimizes expected loss.  
We call such an action $a^*$ an {\em optimal\/} action for $\Pr_Y$.
Suppose that the agent observes the value of a variable
$X$ that takes on values in $\cX$.  Further assume that, although the agent
knows the marginal distribution $\Pr_Y$ of $Y$,
she does not know how $Y$ depends on $X$.  That is, the agent's
uncertainty is characterized by the set $\P$ consisting of all
distributions on $\cX \times \cY$ with marginal distribution $\Pr_Y$ on
$\cY$.  The agent now must choose a {\em decision rule\/} that
determines what she does as a function of her observations.  
We allow decision rules to be randomized.  Thus, if $\Delta(\cA)$ consists
of all probability distributions on $\cA$, a decision rule is a function 
$\delta: \cX \rightarrow \Delta(\cA)$ that chooses a distribution over
actions based on her observations.
Let $\cD(\cX,\cA)$ be the set of all such decision rules.
A special case is a
deterministic decision rule, that assigns probability 1 to a particular
action.  If $\delta$ is deterministic, we 
sometimes abuse
notation and
write $\delta(x)$ for the action that is assigned probability 1 by the
distribution $\delta(x)$.
Given a decision rule $\delta$ and a loss 
function $L$, let $L_\delta$ be the random variable on $\cX \times \cY$
such that $L_\delta(x,y) = \sum_{a \in \cA} \delta(x)(a) L(y,a)$.
Here $\delta(x)(a)$ stands for the probability of performing action $a$
according to the distribution $\delta(x)$ over actions that is adopted
when $x$ is observed.
Note that in 
the special case that $\delta$ is a deterministic decision rule,
then $L_\delta(x,y) = L(y, \delta(x))$.  Moreover, if $\delta_a$ is the
(deterministic) decision rule that always chooses $a$, then 
$L_{\delta_a}(x,y) = L_a(y)$.

The following result, whose proof we leave to the full paper, shows that
the decision rule  
$\delta^*$
that always chooses an
optimal action $a^*$ for $\Pr_Y$, independent of the observation, is
optimal in a minimax sense.
Note that the worst-case expected loss of decision-rule $\delta$ is 
$\sup_{\Pr \in \cP} E_{\Pr}[L_\delta]$.  Thus, the best worst-case
loss 
(i.e., the {\em minimax\/} loss)
over all decision rules is 
$\inf_{\delta \in \cD(\cX,\cA)} \sup_{\Pr \in \cP} E_{\Pr} [L_{\delta}]$.
\begin{proposition}
\label{pro:minimax}
Suppose that $\Pr_Y$ is an arbitrary distribution on $\Y$, $L$ is an
arbitrary loss function, $\cP$ consists of all distributions on $\X
\times \Y$ with marginal $\Pr_Y$, and 
$a^*$ is an optimal action for $\Pr_Y$ (with respect to the loss
function $L$).  Then
$E_{\Pr_Y}[L_{a^*}] = 
\inf_{\delta \in \cD(\cX,\cA)} \sup_{\Pr \in \cP} E_{\Pr} [L_{\delta}]$.
\end{proposition}
\commentout{
\prf
Before doing the proof, we need to introduce some notation.
$L_\delta(x,y)$ is a function of two
variables.  We have been using $E_{\Pr}[L_\delta]$ to denote the
expected value of $L_\delta$ with respect to a distribution $\Pr \in
\cP$ on $\cX \times \cY$.  For each fixed value $x$
of $X$, we can view $L_\delta(x,y)$ as a function of $y$, so it makes
sense to
take its expected value with respect to $\Pr_Y$, which we denote
$E_{\Pr_Y}^Y[L_\delta(x,Y)]$.  
(Note that the superscript in $E_{\Pr_Y}^Y$ denotes the random variable
whose expectation we are taking, and the subscript $\Pr_Y$ denotes the
probability distribution on $Y$ with respect to which the expectation is
taken.  We sometimes omit the superscript or subscript when it is clear
from context.)
We can then take the expectation of this
random variable with respect to a distribution $\Pr_X$ on $\cX$.
Similarly, we can have 
$E_{\Pr_Y}^Y[E_{\Pr_X}^X[L_\delta(X,Y)]]$.
Next, note that it is almost immediate that if $\Pr$ is any distribution
on $\cX \times \cY$, since $L_{\delta_{a^*}}$ does not depend on the
value of $\cX$, we have
\begin{equation}\label{eq1}
E_{\Pr_Y}^Y [L_{a^*}] = E_{\Pr}[L_{\delta_{a^*}}] = \sup_{\Pr' \in \cP}  
E_{\Pr'} [L_{\delta_{a^*}}].
\end{equation}
We now have, 
for every distribution $\Pr_X$ on $X$,
\begin{equation}
\label{eq:deadline}
\begin{array}{ll}
E_{\Pr_Y}^Y[L_{a^*}]
& = \inf_{a \in \cA}   E^Y_{\Pr_Y } [L_a]\\
& = \inf_{\delta \in \cD(\cX,\cA)}  E^X_{\Pr_X} [E^Y_{\Pr_Y } 
[L_\delta(X,Y)]].
\end{array}
\end{equation}
Note that $E^X_{\Pr_X} [E^Y_{\Pr_Y}[L_\delta(X,Y)]] =
E^{(X,Y)}_{\Pr'}[L_\delta(X,Y)]$ for some distribution $\Pr'$
according to which $X$ and $Y$ are independent, with marginals $\Pr_X$
and $\Pr_Y$ respectively. Therefore (\ref{eq:deadline}) can be further
rewritten as
$$
\begin{array}{ll}
E_{\Pr_Y}^Y[L_{a^*}] & = 
\inf_{\delta \in \cD(\cX,\cA)} E^{(X,Y)}_{\Pr'}[L_\delta(X,Y)] \\
& \leq \sup_{\Pr \in \cP} \inf_{\delta \in \cD(\cX,\cA)} E^{(X,Y)}_{\Pr} [L_\delta]\\
& \leq  \inf_{\delta \in \cD(\cX,\cA)} \sup_{\Pr \in \cP} E^{(X,Y)}_{\Pr} [L_\delta]\\
& \leq \sup_{\Pr \in \cP} E^{(X,Y)}_{\Pr} [L_{\delta_{a^*}}]\\
& = E^Y_{\Pr_Y}[L_{a{^*}}] \ \ \ \ \ \ \mbox{[by (\ref{eq1})]}.
\end{array}
$$ \eprf
}
\commentout{
For part (c), 
if the loss function is strict, suppose that $\Pr^*$ be such that $X$
and $Y$ are independent according to $\Pr^*$ and $\Pr$ is such that 
$X$ and $Y$ are not independent.
$$
\sup_{\Pr' \in \cP} E_{\Pr'} [L_{\delta_{\Pr}}] \geq E_{\Pr^*}
[L_{\delta_{\Pr}}]  
> E_{P^*} [L_{a^*}],$$
where the last inequality follows by strictness of the loss function.
\eprf
}

A standard decision rule when uncertainty is represented by a set $\P$ of
probability measures is the Maxmin Expected Utility Rule \cite{GS1989};
compute the expected utility (or expected loss) of an action with
respect to each of the probability measures in $\P$, and then choose the
action whose worst-case expected utility is best (or worst-case expected
loss is least).   Proposition~\ref{pro:minimax} says that if $\P$
consists of all probability measures with marginal $\Pr_Y$ and the loss
depends only on the value of $Y$, then the action with the least
worst-case loss is an optimal action with respect to $\Pr_Y$. 
\commentout{
In words,
\cor
\label{cor:optrel}
If only the marginal distribution $\Pr_Y$ is known, and all
distributions on $\cX \times \cY$ compatible with $\Pr_Y$ are deemed possible,
then adopting $P_Y$ (that is, ignoring $X$) is
\begin{enumerate}
\item minimax optimal (Equation~\ref{eq:ignoptimal})
\item reliable (Equation~\ref{eq:ignreliable})(the expected loss of
  the Bayes act $a^*$ under $P_Y$ is equal to the expected loss of
  $a^*$ under the unknown, ``true'' distribution).
\end{enumerate}
\ecor 
}
\xam
\label{ex:nonbayes1} 
Consider perhaps the simplest case, where
$\cX = \cY = \{0,1\}$. Suppose 
that
our agent knows that $E_{\Pr_Y}[Y] = \Pr_Y(Y = 1) = p$ for
some fixed $p$. As  before, let $\cP$ be the set of distributions 
on $\cX \times \cY$ with 
marginal
$\Pr_Y$.  
Suppose further that the only actions are 0 and 1 (intuitively, these actions
amount to predicting the value of $Y$), and that the loss function is 0
if the right value is predicted and 1 otherwise; that is,
$L(i,j) = |i-j|$. This is the so-called $0/1$ or {\em classification\/} loss.
It is easy to see that 
$E[L_0] = p$ and $E[L_1] = 1-p$, so the optimal act is to choose 0 if $p
< .5$ and 1 if $p > .5$ (both acts have loss $1/2$ if $p = .5$).  
The loss of the optimal act is $\min(p,1-p)$. 

Perhaps the more standard approach for dealing with uncertainty in this
case is to work with the whole set of distributions.  Assume that 
$0 < \Pr(Y=1) = p <1$.  Let $\cP_i = \{\Pr(\cdot \mid X = i): \Pr \in
\cP\}$, $i = 0, 1$.  Then for all $q \in [0,1]$, both $\cP_0$ and
$\cP_1$ contain a 
distribution $\Pr_q$ such that $\Pr_q(Y=1) = q$. In other words, $\cP_0
= \cP_1 = \Delta(\cY)$, the set of all distributions on $\cY$.
Observing $X=x$ causes all information about 
$Y$ to be lost. Remarkably, this holds {\em no matter what value of $X$
  is observed}. 

Thus, even though the agent knew that $\Pr(Y=1) = p$ before observing
$X$, after observing $X$, the agent has no idea of the probability that
$Y=1$.  
This is a special case of a phenomenon that has been
called {\em dilation\/} in the statistical and imprecise probability
literature 
\cite{Augustin03,CozmanWalley,HerronSW97,SeidenfeldW93}:~it is possible
that lower 
probabilities strictly decrease and  that upper probabilities strictly
increase, no matter what value of $x$ is observed. 
Dilation has severe consequences for decision-making.
The minimax-optimal decision rule $\delta^*$ with respect to $\cP_Y$ is to
randomize, choosing both 0 and 1 with probability $1/2$.  
Note that, no matter what $\Pr \in \cP$ actually obtains,
$$
E_{\Pr} [L_{a^*}] = \min \{ p, 1- p \} \ \ ; \ \ E_{\Pr} [L_{\delta^*}] = 1/2.
$$
Thus, if $p$ is close to 0 or 1, ignoring information does much better
than making use of it.

This can be viewed as an example of what decision theorists have called
{\em time inconsistency}.  Suppose, for definiteness, that $p = 1/3$.
Then, a priori, the optimal strategy is to decide 0 no matter what.
On the other hand, if either $X=0$ or $X=1$ is observed, then the
optimal action is to randomize.   When uncertainty is described with a
single probability distribution (and updating is done by conditioning),
then time inconsistency cannot occur.%
\footnote{It has been claimed that this time consistency also depends
on the agent having perfect recall; see \cite{Hal15,PR97} for some
discussion of this issue.}
\exam 
\commentout{
For a given distribution $P$, 
the {\em lower conditional probability of $y$ given
  x\/} is defined as
$$
\underline{P}(Y=y \mid X = x) := \inf_{P \in \cP} \{ P(Y= y \mid X= x) \}. 
$$
The upper probability $\overline{P}(Y=y \mid X = x) $ is defined
analogously.
}
\section{WHEN IGNORING HELPS: A BAYESIAN ANALYSIS}
\label{sec:bayes}
\commentout{
\subsection{The Bayesian case}

Suppose that, instead of having just a set $\cP$ of probability
measures, the agent has a probability measure on $\cP$.  But then which
probability measure should she take?  In the absence of further
information, the standard Bayesian recommendation is to take the uniform
prior.  Of course, exactly what counts as ``uniform'' is not clear when
the state space is continuous.  However, we now show that, for a
reasonable notion of continuous, ignoring information is better than
assuming a uniform distribution on $\cP$.  

Consider perhaps the simplest case, where $\cX = \cY = \{0,1\}$.  For
definiteness, suppose that the known prior $\Pr_Y$ is such that
$\Pr_Y(0) = .3$.  A probability measure on $\cX \times \cY$ is
completely determined by $\Pr(X=0 \mid Y=0)$ and $\Pr(X=0 \mid Y=1)$.
Moreover, for every choice $(\alpha,\beta) \in [0,1] \times [0,1]$ for
these two conditional probabilities, there is a probability
$\Pr_{\alpha,\beta} \in \cP$; in fact 
$$\begin{array}{l}
{\Pr}_{\alpha,\beta}(0,0) = .3 \alpha; \ \ 
{\Pr}_{\alpha,\beta}(0,1) = .7 \beta; \\
{\Pr}_{\alpha,\beta}(1,0) = .3 (1 - \alpha); \ \ 
{\Pr}_{\alpha,\beta}(1,1) = .7 (1 - \beta).
\end{array}
$$
Given this, one obvious way to put a uniform prior on $\cP$ is just to
take a uniform prior on the square $[0,1]^2$.  
Suppose that the only actions are 0 and 1 (intuitively, these actions
amount to predicting the value of $Y$), and that the loss function is 0
if the right value is predicated and 1 otherwise; that is,
$L(i,j) = |i-j|$.  

Notice that $\Pr_{\alpha,\beta}(X=0) = .3 \alpha + .7 \beta$, so
$\Pr_{\alpha,\beta}(Y = 0 \mid X=0) = .3 \alpha/(.3 \alpha + .7 \beta)$.
Thus, the expected loss of predicting 0 if the true probability
distribution is $\Pr_{\alpha,\beta}$ is .$3 \alpha/(.3 \alpha + .7
\beta)$.  Thus, the expected expected loss of predicting 1 is given that
$X=0$ is observed is
$$
\begin{array}{ll}
&\int_{\alpha = 0}^1 \int_{\beta = 0}^1 .3 alpha/(.3 \alpha + .7
\beta) d \alpha d \beta\\
= &\int_{\alpha = 0}^1  (3 \alpha/7)\ln(.3 \alpha + .7 \beta)
|_{\beta=0}^{\beta=1}\\
= &\int_{\alpha = 0}^1 3\alpha/7 \ln((3\alpha+7)/3\alpha) d \alpha\\
= &(3\alpha^2/14 \ln((3\alpha + 7)/3\alpha) + (\alpha/2) - (7/6)\ln(6
\alpha + 14)|_{\alpha=0}^1 \ \ \ \mbox{[see below]}\\
= &1/2 + (3/14)\ln(10/3) - (7/6)\ln(20) + (7/6)\ln(14)\\
= &1/2 + (3/14)\ln(10/3) + (7/6)\ln(.7)
\end{array}
$$
The integral is computed by observing that the derivative of 
$3\alpha^2/14 \ln((3\alpha +7)/3\alpha)$ is
$$\begin{array}{ll}
&\frac{3\alpha}{7}(\ln(\frac{3\alpha+7}{3 \alpha}) +
(\frac{3\alpha^2}{14}\frac{3\alpha}{3\alpha+7} (\frac{3}{3\alpha}
- \frac{3(3\alpha + 7)}{(3\alpha)^2})\\
= &\frac{3\alpha}{7}(\ln(\frac{3\alpha}{3 \alpha + 7}) +
 (\frac{3\alpha^2}{14}\frac{3\alpha}{3\alpha+7} \frac{-21}{9\alpha^2})\\
= &\frac{-3\alpha}{6\alpha+14}\\
= &-\frac{1}{2} + \frac{7}{6\alpha+14}
\end{array}
$$
The $\alpha/2 - (7/6)\ln(6\alpha + 14)$ term in the integral above is
just what is needed to get something whose derivate cancels out
the $-(1/2) + 7/(6\alpha + 14)$.

Similarly,  $\Pr_{\alpha,\beta}(X=1) = .3 (1- \alpha) + .7 (1 -\beta$), so
$\Pr_{\alpha,\beta}(Y = 0 \mid X=1) = .3(1- \alpha/(.3 (1- \alpha + .7
(1-\beta))$. 
Thus, the expected loss of predicting 0 if the true probability
distribution is $\Pr_{\alpha,\beta}$ is $.3 \alpha/(.3 \alpha + .7
\beta)$.  Thus, the expected expected loss of predicting 1 is given that
$X=1$ is observed is also $1/2 + (3/14)\ln(10/3) + (7/6)\ln(.7)$.
This is roughly $.5 + .258 - .416 = .342.$  [[PETER, I'M NOT QUITE SURE
WHAT TO MAKE OF THIS.  I THINK MY CALCULATIONS ARE CORRECT.]]

Notice for future reference that $X$ and $Y$ are independent if 
$\alpha = \beta$.  Thus, the set of distributions that make $X$ and $Y$
independent get probability 0 according to the uniform prior.  Since the
information about the value of $X$ can safely be ignored only if $X$ and
$Y$ are independent, this prior says that, almost surely, $X$ should not
be ignored.  
} 
Suppose that, instead of having just a set $\cP$ of probability
measures, the agent has a probability measure on $\cP$.  But then
which probability measure should she take? 
Broadly speaking, there are two possibilities here.  We can consider
either {\em purely subjective\/} Bayesian agents or {\em pragmatic\/}
Bayesian agents. A purely subjective Bayesian agent 
will come up with some (arbitrary)  prior that
expresses her subjective beliefs about the situation. It then makes
sense to assess the consequences of ignoring information in terms of
expected loss, where the expectation is taken with respect to the
agent's subjective 
prior.  Good's total evidence theorem, a classical result of Bayesian
decision theory \cite{Good67,RaiffaS61}, states 
that, when taking the expectation with respect to the agent's prior, 
the optimal decision should always be based on 
conditioning on {\em all\/} the available information---information
should never be ignored.

In contrast, we consider an agent who adopts Bayesian updating for
pragmatic 
reasons
(i.e., because it usually works well) rather than for
fundamental reasons.  
In this case, because computation time is limited and/or
prior knowledge is hard to obtain or formulate, the prior adopted is
typically easily computable and ``noninformative'', such as a prior
that is uniform in some natural parameterization of $\cP$.  
We suspect that many statisticians are 
pragmatic Bayesians in this sense.
(Indeed,
most ``Bayesian'' UAI and statistics papers adopt pragmatic priors
that cannot seriously be viewed as fully subjective.) 
When
analyzing such a pragmatic approach, it no longer makes that much
sense to compare ignoring information to Bayesian updating on new 
information by looking at the expected loss with respect to the adopted
prior. The reason is that the 
prior can no longer be expected to correctly reflect the agent's
degrees of belief.  It seems more meaningful to pick a single 
probability measure $\Pr$ and to analyze the behavior of the Bayesian
under the assumption that $\Pr$ is the ``true'' state of nature. 
By varying $\Pr$ over the set $\cP$, we can get a 
sense of the behavior of Bayesian updating in all possible
situations. This is the type of analysis that we adopt in this
section;
it is quite standard in the statistical literature
on consistency of Bayes methods \cite{BlackwellD62,Ghosal98}.
We focus on a large class 
of priors on $\cP$ that includes most standard recommendations for
noninformative priors. Essentially, we show that for any prior in the
class, when the sample size is small, ignoring information is better
than using the Bayesian posterior. 
That is, if a pragmatic agent has the choice between (a) first 
adopting a pragmatic prior, 
perhaps not correctly reflecting her own beliefs, and then reasoning 
like a Bayesian, or, (b) simply ignoring the available information, 
then, when the sample size is small, she might prefer option (b). 
On the other hand, as more information becomes available, the Bayesian
posterior
behaves almost as well
as ignoring the information in the worst case, and substantially
better than ignoring in most other cases.
\commentout{In the absence of further
information, the standard Bayesian recommendation is to take the
uniform prior. Of course, exactly what counts as ``uniform'' is not
clear when the state space is continuous.  However, we now show that,
for a reasonable notion of continuous, ignoring information 
can be viewed as being better than assuming a uniform distribution on
$\cP$.}
(Of course, part of the issue here is what counts as ``better''
when uncertainty is represented by a set of probability measures.
For the time being, we say that ``A is better than B'' if A achieves better 
minimax behavior than B.  We return to this issue at the end of this 
section.)

\xam\label{ex:bayes1} 
As in Example~\ref{ex:nonbayes1}, let
$\cX = \cY
= \{0,1\}$.  For definiteness, suppose that the known prior $\Pr_Y$ is
such that 
$\Pr_Y(Y=1) = p$.
Throughout this section we assume that $0 < p < 1$.
A probability measure on $\cX \times \cY$ is
completely determined by $\Pr(X=1 \mid Y=1)$ and $\Pr(X=1 \mid Y=0)$.
Moreover, for every choice $(\alpha,\beta) \in [0,1] \times [0,1]$ for
these two conditional probabilities, there is a probability
$\Pr_{\alpha,\beta} \in \cP$; in fact
$$\begin{array}{c}
{\Pr}_{\alpha,\beta}(X=1, Y=1) = p \alpha; \\
{\Pr}_{\alpha,\beta}(X=1, Y=0) = (1-p) \beta; \\ 
{\Pr}_{\alpha,\beta}(X= 0, Y= 1) = p (1 - \alpha); \\
{\Pr}_{\alpha,\beta}(X= 0, Y=0) = (1-p) (1 - \beta).
\end{array}
$$
Notice that $\Pr_{\alpha,\beta}(X=1) = p \alpha + (1-p) \beta$.
Given this, one obvious way to put a uniform prior on $\cP$ is just to
take a uniform prior on the square $[0,1]^2$; 
we adopt this prior for the time being and consider 
other notions of ``uniform'' further below. 

To calculate the Bayesian predictions of $Y$ given $X$, we must first 
determine the Bayesian 
``marginal'' probability measure
$\overline{\Pr}$, 
where 
$\overline{\Pr}(X=i, Y=j) = \int_{\alpha = 0}^1 \int_{\beta = 0}^1 
\Pr_{\alpha,\beta}(X=i,Y=j) d\alpha d \beta$
(``marginal'' because we are marginalizing out the parameters $\alpha$
and $\beta$),
and then
use $\overline{\Pr}$ to calculate the expected loss of predicting 
$Y=1$. That is,
we calculate 
the so-called ``predictive distribution''
$\overline{\Pr}(Y= \cdot \mid X= \cdot)$.
We can calculate this directly without performing any integration as
follows. By symmetry, we must
have $\overline{\Pr}(Y=1 \mid X= 1) = \overline{\Pr}(Y=1 \mid X= 0)$.
Now if $\gamma = \overline{\Pr}(X=1)$, then
it must be the case that $\gamma\overline{\Pr}(Y=1 \mid X= 1) + (1- 
\gamma)\overline{\Pr}(Y=1 \mid X= 0) = p$; this implies that
$\overline{\Pr}(Y=1 \mid X= 1) = p$. Thus, when calculating the
predictive distribution of $Y$ 
  after observing $X$, the Bayesian will always ignore the value of $X$ and
  predict with his marginal distribution $\overline{\Pr}_Y$.

Thus, before observing data, the Bayesian ignores the value
of $X$, and thus makes minimax-optimal decisions.
Potentially suboptimal behavior of the Bayesian can occur only {\em
  after\/} the Bayesian has observed some data. 
To analyze this case, we need to assume that we
have a sequence of $n$ observations $(X_1, Y_1), \ldots, (X_n,Y_n)$ and
are trying to predict the value of $Y_{n+1}$, given the value of
$X_{n+1}$.  The distribution $\Pr_{\alpha,\beta}$ on $\X \times \Y$ 
is extended to a distribution $\Pr_{\alpha,\beta}^n$ on $(\X \times
\Y)^n$ by assuming that the observations are independent.  Of
course, the hope is that the observations will help us learn about
$\alpha$ and $\beta$,  
allowing us to make better decisions on $Y$.
To take the simplest case, suppose that we have observed that $(X_1,Y_1)
= (1,1)$, and $X_2 = 1$, 
and want to calculate the value of $Y_2$.  Note that
$$\begin{array}{ll}
&\overline{\Pr}^2(Y_2 = 1 \mid X_2 = 1, (X_1,Y_1)= (1,1))\\
= &\frac{\overline{\Pr}^2((X_2,Y_2) = (1,1),(X_1,Y_1)= (1,1)) }{
\mbox{$\sum_{y \in \{0,1\}}$} \overline{\Pr}^2((X_2,Y_2) = (1,y),(X_1,Y_1)=
(1,1)) } \\
= &\frac{\int_{\alpha = 0}^1 \int_{\beta = 0}^1 (p \alpha)^2 d \alpha d \beta}
{\int_{\alpha = 0}^1 \int_{\beta = 0}^1 (p \alpha)^2 d \alpha d \beta 
+ 
\int_{\alpha = 0}^1 \int_{\beta = 0}^1 (p \alpha)(1-p) \beta d \alpha
d \beta}  \\
= & \frac{\frac{1}{3}p}{\frac{1}{3}p + (1-p)\frac{1}{4}}\\
= &\frac{4p}{p+3}. 
\end{array}
$$
Since we must have 
$\overline{\Pr}^2(Y_2 = 1 \mid  (X_1,Y_1)= (1,1)) = p$, it follows using
the same symmetry argument as above that
if $(X_1,Y_1) = (1,1)$, then, no matter what value of $X_2$ is observed,
the value of $X_2$ is not ignored. 
Similar calculations show that, if 
$(X_1,Y_1) = (i,j)$ for all $i, j \in \{0,1\}$, 
 then, no matter what value of $X_2$ is observed,
the value of $X_2$ is not ignored. 
\exam

\commentout{
In light of Proposition~\ref{pro:minimax}(c), this
already shows that the Bayesian
prediction of $Y_2$ given $X_2$ cannot be minimax optimal for all loss
functions.  
We defer discussion of how serious a problem this is until we have seen
a few more examples. 
\commentout{However, we can already point out now why
computing the expectation with respect to $\overline{\Pr}$ is more
appropriate that computing the expected loss over all
$\Pr_{\alpha,\beta}$: Because of independence, $\Pr_{\alpha,\beta}
(Y_2 = i  \mid X_2 = j, (X_1,Y_1)= (i',j')) =
\Pr_{\alpha,\beta}\Pr_{\alpha,\beta}(Y_2 = i \mid X_2 = j)$; similarly
with repeated observations.  That is,
with the first approach, there would be no learning from data.
}
}

In Example~\ref{ex:bayes1} we claimed that the Bayesian should predict
by the predictive distribution $\overline{\Pr}^2(Y = 1 \mid X = 1)$ as
defined in the example. 
While this is the standard Bayesian approach, one may also directly
consider the ``expected'' conditional probability 
$\int_{\alpha = 0}^1 \int_{\beta = 0}^1 
\Pr^2_{\alpha,\beta}(Y=1 \mid X=1)$.  These two approaches give different
answers, since expectation does not commute with division. 
To see why we prefer the standard approach, note that, 
because of independence, $\Pr^2_{\alpha,\beta}
(Y_2 = i  \mid X_2 = j, (X_1,Y_1)= (i',j')) =
\Pr_{\alpha,\beta}
((X_1,Y_1)= (i',j'))
\Pr_{\alpha,\beta}(Y_2 = i \mid X_2 = j)$; similarly
with repeated observations.  That is,
with the alternative approach, there would be no learning from data.
Thus, for the remainder of the paper, we use the predictive-distribution
approach, with no further comment.

\commentout{
We remark that there is no unique way of defining ``uniform prior''.
Another standard approach is to consider the uniform distribution over
the four parameters
$\theta_{xy} = \Pr(X=x,Y=y), x,y \in \{0,1\}$,
restricted to the simplex $\theta_{00} +\theta_{10} = p$, 
$\theta_{01} + \theta_{11} = 1-p$.
The calculations are somewhat more complicated, but lead to
qualitatively similar results (with somewhat larger discrepancies
between Bayes and minimax).  We omit the details here.
}
\xam\label{ex:bayes3}
Now consider the more 
general situation where $\cX = \{1, \ldots, M \}$ for arbitrary $M$, and $
\cY = \{0,1\}$ as before.
We consider a 
straightforward extension of the previous set of distributions: let
$\vec{\alpha} = (
\alpha_1, \ldots, \alpha_M)$ be an element
of the 
$M$-dimensional unit simplex; $\vec{\beta}$ is defined
similarly. 
Fix $p \in [0,1]$, and define
$$
\begin{array}{c}
{\Pr}_{\vec{\alpha},\vec{\beta}}(Y  = 1) = p; \  
{\Pr}_{\vec{\alpha},\vec{\beta}}(X= j \mid Y = 1) = \alpha_j; \\  
{\Pr}_{\vec{\alpha},\vec{\beta}}(X= j \mid Y = 0) = \beta_j.
\end{array}
$$
Note that
$$
{\Pr}_{\vec{\alpha},\vec{\beta}}(X=j,Y=1) = 
\alpha_j p
$$
and
$$
{\Pr}_{\vec{\alpha},\vec{\beta}}(X=j,Y=0) = 
\beta_j (1-p).
$$
Let $D$ be a random variable used to denote the outcome of then $n$
observations $(X_1,Y_1), \ldots, (X_n,Y_n)$.  Given a sequence
$(\vec{x},\vec{y}) = ((x_1,y_1), \ldots, (x_n,y_n))$ of observations,
let
$n^{(\vec{x},\vec{y})}_{k}$ denote the number of observations in the sequence
with $Y_i = k$, for 
$k \in \{0,1\}$.
Similarly, $n^{(\vec{x},\vec{y})}_{(j,k)}$ denotes the number of
observations $(X_i,Y_i)$ in the sequence with 
$(X_i=j,Y_i=k)$.
Then
$$
\begin{array}{ll}
&{\Pr}_{\vec{\alpha},\vec{\beta}}^n(D=(\vec{x},\vec{y}))\\
= &p^{n_1^{(\vec{x},\vec{y})}}(1-p)^{n_0^{(\vec{x},\vec{y})}} \prod_{j=1}^M
\alpha_j^{n_{(j,1)}^{(\vec{x},\vec{y})}}  
\prod_{j=1}^M \beta_j^{n^{(\vec{x},\vec{y})}_{(j,0)}}. 
\end{array}
$$

We next put a 
prior on
$\cP = 
\{\Pr_{\vec{\alpha},\vec{\beta}}: \alpha, \beta \in [0,1]\}$
We restrict attention to priors that can be written as a product of
{\em Dirichlet distributions\/} \cite{BernardoS94}. A
Dirichlet distribution on the $M$-dimensional unit simplex $\Delta_M$
(which we can identify with the set of probability distributions on
$\{1, \ldots, M\}$)
is parameterized by an $M$-dimensional vector $\vec{a}$. For $\vec{a}
= (a_1,
\ldots, a_M)$, the $\vec{a}$-Dirichlet distribution has  density $p_{\vec{a}}$ that satisfies,
for all $\vec{\alpha} \in \Delta_M$,
$$
p_{\vec{a}}(\vec{\alpha}) = \frac{1}{Z(\vec{a})} \alpha_1^{a_1-1}
\cdot \ldots \cdot \alpha_M^{a_M-1},
$$
where $Z(\vec{a}) = \int_{\vec{\alpha} \in \Delta_M} \alpha_1^{a_1-1}
\cdot \ldots \cdot \alpha_M^{a_M-1} d \vec{\alpha} $
is a normalizing factor.
Note that the uniform prior is the $\vec{a}$-Dirichlet prior
where $a_1 = a_2 = \ldots = a _M = 1$.  As we shall see, 
many other priors of interest are special cases of Dirichlet priors.

We consider only priors $w$ on $\cP$
that satisfy $w(\vec{\alpha},\vec{\beta}) = w_{\vec{a}}(\vec{\alpha}) 
w_{\vec{b}}(\vec{\beta}) $ for all $\vec{\alpha}, \vec{\beta} \in \Delta_M$,
where $w_{\vec{a}}$ and $w_{\vec{b}}$ are of the Dirichlet form.
Then 
\begin{equation}
\label{eq:dirichlet}
\begin{array}{ll}
&\overline{\Pr}^n(D = (\vec{x},\vec{y}))\\ 
= &\int_{\vec{\alpha} \in \Delta_M} \int_{\vec{\beta} \in \Delta_M}
\mbox{$\Pr$}^n_{\vec{\alpha},\vec{\beta}}(D = (\vec{x},\vec{y})) 
w_{\vec{a}}(\vec{\alpha}) w_{\vec{b}}(\vec{\beta})
d \vec{\alpha} d \vec{\beta}.
\end{array}
\end{equation}

Now suppose that a Bayesian has observed an initial sample $D$ of size $n$
and $X_{n+1}$, and must predict $Y_{n+1}$.
Suppose $X_{n+1} = k$. Then the Bayesian's predictive distribution
becomes $\overline{\Pr}^{n+1}(Y_{n+1} = \cdot \mid X_{n+1}, D)$ or, 
more explicitly, 
$$\begin{array}{ll}
&\overline{\Pr}^{n+1}(Y_{n+1} = j \mid X_{n+1} = k, D=(\vec{x},\vec{y})) \\
= &\frac{\overline{\Pr}^{n+1}(D = (\vec{x},\vec{y}), X_{n+1} = k, Y_{n+1} =
j)}{\overline{\Pr}^{n+1}(D = (\vec{x},\vec{y}), 
  X_{n+1} = k)}. 
\end{array}
$$
It will be
convenient to represent this distribution by the odds ratio 
\begin{equation}
\begin{array}{ll}
\label{eq:oddsratio1}
&\frac{\overline{\Pr}^{n+1}(Y_{n+1} = 1 \mid X_{n+1} = k,
  D=(\vec{x},\vec{y}))}{\overline{\Pr}^{n+1}(Y_{n+1} = 0 \mid 
  X_{n+1} = k, D=(\vec{x},\vec{y}))}\\ 
= 
&\frac{\overline{\Pr}^{n+1}(D= (\vec{x},\vec{y}), X_{n+1} = k, Y_{n+1}
  = 1)}{\overline{\Pr}^{n+1}(D = (\vec{x},\vec{y}),
X_{n+1} = k, Y_{n+1} = 0)}. 
\end{array}
\end{equation}
\commentout{
\nonumber \\
& = &
\frac{p^{n_1^y + 1}(1-p)^{n_0^y} \ \cdot \ 
\int \alpha_k \prod_{j=1}^M \alpha_j^{n_{(j,1)}}
d \vec{\alpha} \ \cdot \ \int  \prod_{j=0}^M \beta_j^{n_{(j,1)}}
d\vec{\beta}}{p^{n_0^y}(1-p)^{n_1^y +1} \ \cdot \ \int \prod_{j=0}^M \alpha_j^{n_{(j,0)}}
d \vec{\alpha} \ \cdot \ 
\int \beta_k \prod_{j=0}^M \beta_j^{n_{(j,1)}}
d\vec{\beta}} \nonumber \\
 \label{eq:dirichlet}
& = & \frac{p}{1-p} \cdot 
\frac{\int \alpha_k \prod_{j=0}^M \alpha_j^{n_{(j,0)}}
d \vec{\alpha}}{\int \prod_{j=0}^M \alpha_j^{n_{(j,0)}}
d \vec{\alpha}}
\cdot
\frac{\int \prod_{j=0}^M \beta_j^{n_{(j,1)}}
d\vec{\beta}}{\int \beta_k \prod_{j=0}^M \beta_j^{n_{(j,1)}}
d\vec{\beta}}.
\end{eqnarray}
}
Both the numerator and the
denominator 
of the right-hand side of (\ref{eq:oddsratio1})
are of the form (\ref{eq:dirichlet}), so 
this expression
is a ratio of Dirichlet integrals. These can be calculated explicitly
\cite{BernardoS94}, giving
\commentout{Now comes a trick: we recognize the second factor in
(\ref{eq:dirichlet}) as the conditional probability of observing
$X=k$ for a
Bayesian who uses a uniform prior for the multinomial distribution on
$\{1, \ldots, M\}$ and has already observed a sample of size $n^y_0$ 
with, for $j = 1..M$, 
$n_{(j,0)}$ occurrences of $j$. But this conditional probability is
just given by {\em Laplace's rule of succession}, and known to be
equal to 
$$
\frac{\int \alpha_k \prod_{j=0}^M \alpha_j^{n_{(j,0)}}
d \vec{\alpha}}{\int \prod_{j=0}^M \alpha_j^{n_{(j,0)}}
d \vec{\alpha}}
 = \frac{n_{(k,0)}+1}{n_0^y + M+1}.
$$
(We could also have computed this directly by performing two complicated
beta-integrals.)  

Reasoning entirely analogously, we can compute that 
$$
\frac{\int \prod_{j=0}^M \beta_j^{n_{(j,1)}}
d\vec{\beta}}{\int \beta_k \prod_{j=0}^M \beta_j^{n_{(j,1)}}
d\vec{\beta}} = \frac{n_1^y + M+1}{n_{(k,1)}+1},
$$
so that the odds-ratio becomes
}
\begin{equation}
\label{eq:oddsratio2}
\begin{array}{ll}
&\frac{\overline{\Pr}^{n+1}(Y_{n+1} = 1 
\mid X_{n+1} = k, D=(\vec{x},\vec{y}))}{\overline{\Pr}^{n+1}(Y_{n+1} = 0 \mid
  X_{n+1} = k, D=(\vec{x},\vec{y}))} \\
= &\frac{\overline{\Pr}^{n+1}(D = (\vec{x},\vec{y}), X_{n+1} = k,
  Y_{n+1} = 1)}{\overline{\Pr}^{n+1}(D = (\vec{x},\vec{y}),
X_{n+1} = k, Y_{n+1} = 0)} \\
= &
\frac{p}{1-p} \cdot
\frac{n^{(\vec{x},\vec{y})}_{(k,1)}+a_k}{n^{(\vec{x},\vec{y})}_{(k,0)}+b_k}
\cdot  
\frac{n^{(\vec{x},\vec{y})}_0 + \sum_{k=1}^M b_k}{n^{(\vec{x},\vec{y})}_1+ \sum_{k=1}^M a_k}.
\end{array}
\end{equation}
With the uniform prior, (\ref{eq:oddsratio2}) simplifies to
\begin{equation}
\label{eq:uniform}
\frac{p}{1-p} \cdot
\frac{n^{(\vec{x},\vec{y})}_{(k,1)}+1}{n_{(k,0)}^{(\vec{x},\vec{y})} +1} \cdot 
\frac{n_0^{(\vec{x},\vec{y})} + M}{n_1^{(\vec{x},\vec{y})}+ M}.
\end{equation}
(\ref{eq:oddsratio2}) and (\ref{eq:uniform}) show that
the odds-ratio behaves like $p/(1-p)$ (which would be the
odds-ratio obtained by ignoring the values of $X$) times
some ``correction factor''. Ideally this correction
factor would be close to $1$ for small samples and then smoothly change
``in the right direction'', so that the Bayesian's predictions are never
much worse than the minimax predictions and, as more data comes in, get
monotonically better and better. 
We now consider two examples to show the extent to which this happens.

\commentout{
Let  $\Pr$ be a distribution such that  $\Pr(Y_i = 1)= p$ and
$(X_i,Y_i)$ are i.i.d. (independent and identically distributed)
according to $\Pr$.  
We consider 
two
choices of $\Pr$ and see 
how the odds-ratio (\ref{eq:oddsratio2}) behaves under them. 
We use the uniform prior in all our examples.
We then relate this behavior to loss functions.
}

First, take $M=2$, and let $\Pr$ be such that 
$\Pr(Y = 1) = p$,  $\Pr(X=1 | Y=1) = 1$,  
and $\Pr(X=0 | Y=0) = 1$. Then, for $k=1$,
(\ref{eq:oddsratio2}) becomes
$$
\begin{array}{ll}
&\frac{\overline{\Pr}^{n+1}(Y_{n+1} = 1 \mid X_{n+1} = 1,
  D=(\vec{x},\vec{y}))}{\overline{\Pr}^{n+1}(Y_{n+1} = 0 \mid 
  X_{n+1} = 1, D=(\vec{x},\vec{y}))}  \\
= &\frac{p}{1-p} \cdot \frac{n^{(\vec{x},\vec{y})}_1+1}{1} \cdot 
\frac{n_0^{(\vec{x},\vec{y})} + 2}{n_1^{(\vec{x},\vec{y})}+2}.
\end{array}
$$
For all but the smallest $n$, with high $\Pr$-probability,
$n_1^{(\vec{x},\vec{y})} \approx pn$.
Thus, the odds ratio tends to infinity, as expected. 
\commentout{
Next, 
suppose that 
$n \in \cS = \{50, 51, \ldots, 100 \}$) 
and that $M \gg n$ (say,
$M=100,000$). 
This type of situation is typical in problems as that discussed in the
introduction.  In a big city, there will typically be many possible
addresses, but the number of patients the doctor has seen is
relatively
small.
Now 
let $\Pr$ be such that, again, $(X_i,Y_i)$, $i = 1,
\ldots, n$ are i.i.d. according to $\Pr$ and $\Pr(Y_i = 1) = p$,  but now
$X$ is uniformly distributed 
(i.e., $\Pr(X = j) = 1/M$).  
For simplicity, assume that $pM$ is an integer. 
Let $\Pr(Y=1 \mid X = j) = 1$ for $j \in \{1, \ldots, pM -1\}$ and
 $\Pr(Y=1 \mid X = j) = 0$ for $j \in \{pM , \ldots, M \}$. 
Since $M \gg n$, we then have that,
for most of values of $k$, $n_{(k,0)} = n_{(k,1)} = 0$, 
in which case 
the second factor in (\ref{eq:uniform}) becomes equal to
$1$.
The third factor is also approximately equal to $1$.
Thus, for most values of $k$,
Bayes behaves approximately like minimax, 
under these assumptions.  However, it
does not behave {\em exactly\/} like minimax.
Indeed,
by choosing the loss function
appropriately, we can make it behave much worse than minimax.%

Pick some small $\delta > 0$ and 
consider a loss function with asymmetric misclassification 
costs, given by
$
L(0,0) = L(1,1) = 0$; $L(0,1) = p$; $L(1,0) = 1- p - \delta$.
The expected loss given the choice of 0 (with respect to the prior
probability $\Pr_Y$ is $E_{\Pr_Y} [L_0] = p(1-p- \delta)$; 
the expected loss given the choice of 1 is 
$E_{\Pr_Y} [L_1] = p(1-p)$. Thus, the optimal action with respect to the
prior $\Pr_Y$ is to choose 0, and thus the optimal minimax action is to
always choose 0.
Now consider the predictions of a Bayesian who uses the uniform prior.
By the law of large numbers, with very high $\Pr$-probability,
for all $n \in \cS$, the third factor in (\ref{eq:uniform}) is
greater than $1+ \epsilon$ for some small $\epsilon > 0$. 
On the other hand, whenever a value $k$ of $X$ is observed for the
first time, the second factor is equal to $1$. 
Since we have assumed that $M \gg \max \cS$, 
with high $\Pr$-probability, for all
$i \neq j, i, j \in \cS$, we have $x_i \neq x_j$. Therefore, with high
$\Pr$-probability, for all $n \in \cS$, 
the Bayesian posterior odds (\ref{eq:uniform}) are
greater than 
$\frac{p}{1-p} (1 + \epsilon)$.
Thus, for some $\epsilon' > 0$, 
$$
\begin{array}{ll}
&\overline{\Pr}(Y_n \\
= &1 \mid X_n = k, (X_1,Y_1) = (x_1,y_1), \ldots,
(X_{n-1},Y_{n-1})\\ 
= &(x_{n-1},y_{n-1}))\\
> &p + \epsilon'.
\end{array}
$$
This means that, according to the Bayesian agent, the expected loss of
action $a \in \{0,1\}$ is 
$$
E_{\overline{\Pr}}\delta [L_a] = (p+ \epsilon')(1-p- \delta)(1-a) + (1-p
- \epsilon')) p a. 
$$
By choosing $\delta$ appropriately, 
this is minimized by $a=1$. 
Thus, for such a $\delta$, the Bayesian will select, with high
$\Pr$-probability, 
the action $a=1$ for all $n \in \cS$. On the other hand,
with high $\Pr$-probability, about a fraction of $1-p$ of $(x_i,y_i)$ with
$i \in \cS$ will have $y_i = 0$. Therefore, 
choosing 0 (the decision made by an agent who ignore the information)
is actually a better choice in all these cases, with
the difference in performance
being, with high $\Pr$-probability, approximately
$$\begin{array}{ll}
&|\cS|(E_{\Pr_Y}[L_1] - E_{\Pr_Y}[L_0)\\
=  &|\cS|((1-p)  p  -  p(1-p-  \delta))\\
 = &|\cS|p\delta. 
\end{array}$$
As $n$ grows larger, some values of $X$ will be observed more than
once and learning will take place. 
Then taking the information into account is better than ignoring it.
}
In the previous example, $X$ and $Y$ were completely correlated.
Suppose that they are independent.  That is, suppose again that $M=2$,
but that $\Pr$ is such that $\Pr(Y=1 \mid X=k) = p$, for $k =0,1$.
For simplicity, we further suppose that $p=1/2$ 
and that $\Pr(X = 0) = \Pr(X = 1) = 1/2$; the same argument applies with
little change if we drop these assumptions.

\commentout{
Now let $\Pr$ be such that, again, 
$\Pr(Y_i = 1) = p$, but now $M=4$, and 
$\Pr(Y=0 \mid X = 0) = \Pr(Y= 1 \mid X = 1) = 1$, $\Pr(Y =
1 \mid X = 2) = \Pr(Y = 1 \mid X = 3) = p$,  $\Pr(Y=0 \mid X = 0) =
\Pr(Y= 1 \mid X = 1) = 1$, $\Pr(Y = 1 \mid X = 2) = \Pr(Y=1 \mid X=3) = p$.  
Thus, learning that $X \in \{2,3\}$ gives no information about $Y$, but
learning that $X \in \{0,1\}$ gives a great deal of information about $Y$.
If $p = 1/2$, then the first factor in (\ref{eq:uniform}) is
equal to $1$. For all but the smallest $n$, with high probability, the
second and third factors are also approximately equal to $1$, so that
Bayes behaves approximately like minimax, under these assumptions.
However, it does not behave {\em exactly\/} like minimax.  Indeed, by
choosing the loss function
appropriately, we can make it behave worse than minimax.
}
Given $\alpha > 1$, consider a loss function $L_\alpha$ with asymmetric
misclassification costs, given 
by $L_\alpha(0,0) = L_\alpha(1,1) = 0$; $L_\alpha(1,0) = 1$; $L_\alpha(0,1) =
\alpha$.  
Clearly,  $E_{\Pr_Y} [L_0] = 0.5$ and 
$E_{\Pr_Y} [L_1] = 0.5\alpha$. Thus, the optimal action with respect to
the prior $\Pr_Y$ is to predict 0, 
and the minimax-optimal action is to always predict 0.
Moreover, the expected loss of predicting 1 is $.5(\alpha - 1)$.

\commentout{
\begin{lemma}
\label{thm:ignore}
Let $\alpha = \frac{n_{\max}+ 2}{n_{\max} + 1}$ and let $q = \Pr(X_i \in
\{2,3\})$. 
Then for all $q \in (0,1]$, there exists a sequence $r_1, r_2,
\ldots$ with $\lim_{h \rightarrow \infty} r_h \geq 1/8$, such that for
all $n_{\max}$ and all $n < n_{\max}$,
$$
E_{\Pr^{n+1}} [L_{a_{D,X_{n+1}}} - L_0]]
\geq 
\frac{1}{2} r_n (q \alpha -1).
$$
\end{lemma}
\prf
}

Now consider the predictions of a Bayesian who uses the uniform prior.  
The Bayesian will predict 1 iff
$$
\begin{array}{ll}
&\frac{E_{\overline{\Pr}(Y_{n+1} 
\mid X_{n+1},D)}[L_1]}{E_{\overline{\Pr}(Y_{n+1} \mid X_{n+1},D)}[L_0]} \\
= &\frac{\alpha \overline{\Pr}(Y_{n+1} = 0 \mid
X_{n+1}=k,D=(\vec{x},\vec{y}))}{\overline{\Pr}(Y_{n+1} 
=1 \mid X_{n+1}=k,D=(\vec{x},\vec{y}))} < 1.
\end{array}
$$
{F}rom the odds-ratio 
(\ref{eq:uniform}) we see that this holds iff
\begin{equation}
\label{eq:badcond}
\begin{array}{ll}
\alpha &< \frac{\overline{\Pr}(Y_{n+1} = 1 \mid X_{n+1} = k ,D=
(\vec{x},\vec{y}))}{\overline{\Pr}(Y_{n+1} 
= 0 \mid X_{n+1} = k ,D = (\vec{x},\vec{y}))}\\
= &
\frac{n^{(\vec{x},\vec{y})}_{(k,1)}+1}{n^{(\vec{x},\vec{y})}_{(k,0)}+1}
\cdot  
\frac{n_0^{(\vec{x},\vec{y})} + 2}{n_1^{(\vec{x},\vec{y})}+ 2}.
\end{array}
\end{equation}
\commentout{
If it holds, then $ L_{a_{D,X_{n+1}}} - L_0]
(b) with probability $1-q$, $X_{n+1} \in \{0, 1\}$. In that case,
the loss
incurred by the Bayesian prediction is still $
\geq 0$. Therefore, if (\ref{eq:badcond}) holds, then
$$
E_{(X_{n+1},Y_{n+1}) \sim \Pr} [\Lbayes - L_0]
 \geq (1-q) \cdot 0 + q \bigl(\frac{1}{2}\alpha - \frac{1}{2}\bigr).
$$
whereas if (\ref{eq:badcond}) does not hold, then Bayes makes the same
decision as minimax and
$E_{(X_{n+1},Y_{n+1}) \sim \Pr}
[\Lbayes - L_0]$  $= 0$. Letting $r_n$ denote the probability, according to the true distribution
$\Pr$ on $(X_1,Y_1), \ldots, (X_n,Y_n)$, that (\ref{eq:badcond})
holds, this gives
\begin{equation}
\label{eq:almost}
\begin{array}{ll}
& E_{D \sim \Pr} E_{(X_{n+1},Y_{n+1}) \sim \Pr} [\Lbayes - L_0] \geq \\
&  r_n \frac{1}{2}(q \alpha -1) + (1- r_n) \cdot 0 =  \frac{1}{2} r_n (q \alpha -1).
\end{array}
\end{equation}
It remains to bound $r_n$.
}
If $\beta$ is the probability 
(with respect to $\Pr^n$) of
(\ref{eq:badcond}), then the 
difference between the Bayesian's expected loss and the expected loss
of someone who ignores the information is
$\beta(\alpha-1)/2$.  Clearly, $\beta$ depends on $\alpha$ and $n$.
Moreover, for any fixed $\alpha > 1$, $\lim_{n\rightarrow \infty}  \beta
\rightarrow 0$.  This, of course, just says that eventually the Bayesian
will learn correctly.  However, for relatively small $n$, it is not hard to
construct situations where $\beta(\alpha - 1)/2$ can be nontrivial.
For example, if $n=4$ and $\alpha = 1.4$, then $\beta \sim .35$. 
(We computed this by a brute force calculation, by considering all
the values $n_{(k,j)}^{(\vec{x},\vec{y})}$ that cause (\ref{eq:badcond})
to be true, and computing their probability.)  Thus, the Bayesian's
expected loss is about 14\% worse than that of an agent who ignores the
information. 
\commentout{
(\ref{eq:critical}) 
that a sufficient condition for
(\ref{eq:badcond}) to hold is if 
\begin{equation}
\label{eq:badb}
n_0^{(\vec{x},\vec{y})} \geq n_1^{(\vec{x},\vec{y})} \mbox{\ and \ } n_{(k,1)} > 
n_{(k,0)}.
\end{equation}
It is straightforward to show (using the law of large numbers and the
fact that $p = 1/2$) that the
probability of obtaining a sample $D$ satisfying (\ref{eq:badb})
converges to some value $r' \geq 1/8$, the exact value of $r'$ depending
on $q$. Since $r_n$ must converge to  some $r^* \geq r'$, the theorem follows.
\eprf

Lemma~\ref{thm:ignore} says that a Bayesian agent's expected
loss for the ($n+1$)-st outcome, based on the first  
$n$ outcomes, is at least $0.5 r_n (q\alpha -1)$ greater than that of an 
agent that simply ignores the data.
By choosing $q$ close enough to $1$ (but not necessarily equal to
$1$), this can be made strictly larger than $0$,
so that the expected loss of the Bayesian posterior
prediction after having seen a sample of size $n$ is larger than the
expected loss obtained by ignoring the $X_i$. 
If $n$ grows much larger than $n_{\max}$, Bayes will have learned the
distribution $\Pr$ well enough so that it outperforms prediction by
ignoring unless $q=1$.  If $q=1$, then ignoring
is the optimal strategy against the true probability, and the expected
performance of ignoring will be slightly better than that of Bayes for
all values of $n$. 
\exam
\commentout{
This example shows that {\em even if there are strong
  dependencies }between $X$ and $Y$, for small samples, ignoring the
  value of $X$ is sometimes preferable to predicting with the
  Bayesian strategy.
}
The example shows that, even if there is some dependency
between $X$ and $Y$, for small samples, ignoring the value of $X$ is
sometimes preferable to predicting with the Bayesian strategy. The
maximum sample size at which ignoring outperforms Bayes can be made
larger by considering a larger set of observations ${\cal X}$, so that
$n$ must be larger before observations get observed repeatedly, and
Bayes remains ``confused'' for a longer time.
}

Although in this example there is no dependence between $X$ and $Y$ in
the actual distribution, by continuity, the same result holds if there
is some dependence.
\exam

This conclusion assumed that a Bayesian chose a particular 
noninformative
prior, but it does not depend strongly on this choice.
As is well known, 
there is no unique way of defining a ``uniform prior'' on a set of
distributions $\cP$, since what is ``uniform'' depends on the chosen
parameterization. For this reason, people have developed other types
of noninformative priors. One of the most well-known of these is the
so-called Jeffreys' prior \cite{Jeffreys46,BernardoS94}, specifically
designed as a prior expressing ``ignorance''. This prior is
invariant under continuous 1-to-1 reparameterizations of $\cP$.
It turns out that Jeffreys' prior on the set
$\cP$ is also of the Dirichlet form (with $a_1 = \ldots = a_n = b_1 =
\ldots = b_n = 1/2$) so that it 
satisfies (\ref{eq:oddsratio2})
(see, for example, \cite{KontkanenMSTG00}).  
Other pragmatic priors that are often used in practice are the
so-called {\em equivalent sample size (ESS) priors \/}
\cite{KontkanenMSTG00}. For the case of our $\cP$, these also take the
Dirichlet form.  Thus, the analysis of Example~\ref{ex:bayes3} does not
substantially change if 
we use the Jeffreys' prior or an ESS prior.  It remains the case
that, for certain sample sizes, ignoring information is preferable to
using the Bayesian posterior.

Example~\ref{ex:bayes3} shows that with noninformative priors,
for small sample sizes, ignoring the information may be better than
Bayesian updating. 
Essentially, the reason for this is that all standard noninformative
priors assign probability zero to the set of distributions $\cP'
\subseteq \cP$ according to which $X$ and $Y$ are independent.  But the
measures in $\cP'$ are exactly 
the ones that lead to minimax-optimal decisions. 
Of course, there is no reason that a Bayesian must use a noninformative prior.
In some settings it may be preferable to adopt a
``hierarchical pragmatic prior'' that puts a uniform probability on both
$\cP- \cP'$ and $\P'$, and assigns probability $0.5$ to each of $\cP -
cP'$ and $\cP'$. 
Such a prior makes it easier for a Bayesian to learn that $X$ and $Y$
are independent.
(A closely related 
prior has been used by Barron, Rissanen, and Yu  \citeyear{BarronRY98},
in the context of universal coding, with a logarithmic loss function.)
With such a prior, a Bayesian would do better in this example.

The notion of optimality that we have used up to now is 
minimax
loss
optimality.  Prediction $i$ is better than prediction $j$ if the
worst-case loss when predicting $i$ (taken over all possible priors in
$\P$) is better than the worst-case loss when predicting $j$.
But there are certainly other quite reasonable criteria that could be
used when comparing predictions.  In particular, we could consider
minimax regret.  That is, we could consider the prediction that
minimizes the worst-case difference between the best prediction for each
$\Pr \in \P$ and the actual prediction.  In the second half of
Example~\ref{ex:bayes3}, we calculated that if the true probability
$\Pr$ is such that $\Pr(X=0) = \Pr(X=1) =
1/2$ and $\Pr$ makes $X$ and $Y$ independent, then the difference
between the loss incurred by an agent that ignores the prior and a
Bayesian is roughly .07.  We do not know if there are probabilities $\Pr'$
for which the Bayesian agent does much worse than an agent who
ignores the prior with respect to $\Pr'$.  On the other hand, if $X$
and $Y$ are completely correlated, that is, if the
true probability  $\Pr$ is such that $\Pr(Y=1 \mid X=1) = \Pr(Y=0 \mid
X=0) = 1$, then if $n=4$ and $\alpha = 1.4$, the Bayesian will predict
correctly, while half the time the agent that ignores information will
not.   Then the difference between the loss incurred by the Bayesian
agent and the agent that ignores the information is 0.5.  Thus, in the
sense of expected regret, the Bayesian approach is bound to be at least
as good as ignoring the information in this example.  
We are currently investigating whether this is true more generally.

\commentout{Our example only works for weak dependencies between $X$ and $Y$ ($q$
needs to be close to $1$). Since
we used various crude approximations in our calculations, it may be
the case that ignoring can outperform Bayesian inference also if
the dependencies between $X$ and $Y$ are strong ($q$ close to
$0$). We have not been able to establish whether this really is the
case or not.
}
\commentout{
\commentout{
$\theta_{xy} = \Pr(X=x,Y=y), x,y \in \{0,1\}$,
$\theta_{01} + \theta_{11} = 1-p$.
The calculations are somewhat more complicated, but lead to
qualitatively similar results (with somewhat larger discrepancies
between Bayes and minimax).  We omit the details here.
}
TODO SHOULD WE ADD THAT, IF Q EQUAL TO $P^*$ (X AND Y INDEPENDENT),
THEN Bayes SLIGHTLY WORSE THAN MINIMAX EVEN FOR LARGE SAMPLES?
\paragraph{A different type of pragmatic prior}
The analysis following Example~\ref{ex:bayes3} suggests that it may be
a good idea to use a pragmatic prior that is somewhat different from
the usual noninformative priors, which all put $0$ mass on the
ignorance-distribution $P^*$. Instead, let
$(\vec{\alpha},\vec{\beta})^*$ be the parameter vector corresponding
to $P^*$ and consider the hierarchical prior that puts probability
mass $1/2$ on $(\vec{\alpha},\vec{\beta})^*$ and density 
$w'(\vec{\alpha},\vec{\beta})/2$ on $(\vec{\alpha},\vec{\beta}) \neq
(\vec{\alpha},\vec{\beta})^*$. We were not able to show formally that
such a prior is in any sense ``better'' than the uniform or Jeffreys'
prior for classification loss functions of the type used in
Examples~\ref{ex:nonbayes1} and~\ref{ex:bayes3}. Yet the following
example strongly suggests that for the {\em logarithmic\/} loss
function, it is definitely a good idea to use such a prior.  \xam
\label{ex:bayes4}
TODO JOE, I WILL SUPPLY THIS AT A LATER STAGE - FOR THE LOG LOSS IT IS
EASY TO GIVE QUANTITATIVE BOUNDS ON HOW MUCH $P^*$ OUTPERFORMS Bayes IN
THE WORST-CASE ... NO TIME NOW.
\exam
We note that priors of this hierarchical type 
are frequently used as a basis for data compression, for
pragmatic reasons similar to ours  \cite{BarronRY98}.
}
\commentout{
\paragraph{Example 1(c)}:
Now we provide an example in which Bayes behaves better than 1(b)
under a neutral bookie but much worse than 1(b) 
under a semi-adversarial bookie,
(I'm using my earlier terminology here).
Let $Q(X=0 \mid Y= 0) = 1$, $Q(X= j \mid Y=1) = 1/M$ for $j \in
\{1, \ldots, M\}$. Suppose first we observe $X_{n+1} = 0$ (this will
happen with probability $p$). Plugging in
$k = 0$ in (\ref{eq:oddsratio2}), we get
$$
\frac{\overline{\Pr}(Y_{n+1} = 0 \mid X_{n+1} = 0, D)}{\overline{\Pr}(Y_{n+1} = 1 \mid
  X_{n+1} = 0, D)}  = 
\frac{p}{1-p} \cdot \frac{n^{(\vec{x},\vec{y})}_0+1}{1} \cdot 
\frac{n_1^{(\vec{x},\vec{y})} + M+1}{n_0^{(\vec{x},\vec{y})}+M+1},
$$
so that 
$$\overline{\Pr}(Y_{n+1} = 0 \mid X_{n+1} = 0, D) \mbox{\ close\ to\ } 1,$$ 
just as in Example 1(a).

Now suppose we observe $X_{n+1} = k$ for some $k \neq 0, k \neq x_i$
for $i \in \{ 1, \ldots, n\}$ (this will happen with probability
approximately $1-p$). Then we are in the situation of Example 1(b) and
we get
$$ \overline{\Pr}(Y_{n+1} = 0 \mid X_{n+1} = k, D) \approx p.$$
Note that in this case, at least under ``reasonable'' betting schemes,  
Bayes may perform better than the minimax strategy: with
probability $p$, we observe $X_{n+1} = 0$. We then predict $Y_{n+1} =
1$ with probability close to $1$, and $Y_{n+1}$ will indeed be equal
to $1$, so conditional on $X_{n+1} = 0$ we predict better than
minimax. With probability $1-p$ we observe a value of 
$X_{n+1} = k$ we typically 
have not seen before; then $Y_{n+1}$ turns out to be
$1$ and we predict $Y_{n+1} = 1$ with probability close to $p$, which
is equal to what we would have predicted if we had adopted the 
minimax strategy.

Thus, using my earlier terminology, 
if the bookie is {\em neutral}, then (for the given sample size
$n$) Bayes is ``better'' than minimax. Yet if the bookie is {\em
  semi-adversarial}, then Bayesian predictions may be problematic
here - take, for example, $p=0.7$. Then $30\%$ of the time (namely,
whenever $Y=1$) the bookie may offer a ticket which pays $1$ dollar if
$Y=0$, and costs $0.65$ cent. Then the Bayesian will be tempted to buy
such a ticket and so will loose $0.65$ dollar! - note that this will
happen $30\%$ of the time, so it's quite bad. 
\paragraph{Example 1(d)}
It would be nice to have an example of Bayes suffering a lot  compared
to minimax (for example, as much as in Example (c)), but 
with a neutral bookie and bets with 
pay-offs between $0$ and $1$ (so that discrepancies between Bayes and
minimax may not be ``blown up'' by raising the stakes). I couldn't find
such an example. Maybe, such an example doesn't exist? (so Bayes
behaves better than I thought?) Not clear...
\exam
\paragraph{Remark}
I found that one can also do these calculations for the Jeffreys' prior
(very often used as a ``prior expressing ignorance'', invariant under
reparameterization !!!), and the results will be just about the same!

Upon closer inspection, (\ref{eq:oddsratio2}) is just the likelihood
ratio for the so-called naive Bayes (statistical) model with uniform
prior; we should mention this. Since this is well-known, we don't have
to put any of the integral calculations in the paper. Maybe also compare
this to Naive Bayes/take more and more 
features into account as you get more data?
}
\section{PARTIAL IGNORANCE AND DIFFERENT TYPES OF BOOKIES}
\label{sec:bookies}
In Sections~\ref{sec:nonbayes} and~\ref{sec:bayes}, we showed that
ignoring information is sensible as long as 
(1)
the set $\cP$ contains {\em
all\/} distributions on $\cX \times \cY$ with the given marginal $\Pr_Y$, and
(2) the loss function $L$ is fixed; in particular, it does not depend
on the realized value of $X$. 
In this section, we consider what happens when we drop these
assumptions.  

The assumption that $\cP$ contains all distributions on $\cX  \times
\cY$ with the given marginal $\Pr_Y$ amounts to the assumption that all
the agent knows is $\Pr_Y$.  If an agent has more information about the
probability distribution on $\cX \times \cY$, 
then ignoring information is in general not a reasonable thing to do.
To take a simple example, suppose that the set $\P$ contains only
one distribution $P^\circ$. 
Then clearly the minimax optimal strategy is to use
the decision rule based on the conditional distribution $P^\circ(Y \mid
X)$, which means that all available information is taken into account.
Using $P^\circ_Y$ is clearly not the right thing to do.
\commentout{
Similarly, if 
the loss function can depend on the realized value of $X$, 
information can be catastrophic in general. Namely, if the same
scenario (observing $X$, making a decision about $Y$) is repeated
several times and each time a different loss function is used, chosen
by some adversary, then TODO

It turns out that there is an interesting interplay
between both conditions.
TODO
\paragraph{Example 1(c)}:
Now we provide an example in which Bayes behaves better than 1(b)
under a neutral bookie but much worse than 1(b) 
under a semi-adversarial bookie,
(I'm using my earlier terminology here).
Let $Q(X=0 \mid Y= 0) = 1$, $Q(X= j \mid Y=1) = 1/M$ for $j \in
\{1, \ldots, M\}$. Suppose first we observe $X_{n+1} = 0$ (this will
happen with probability $p$). Plugging in
$k = 0$ in (\ref{eq:oddsratio2}), we get
$$
\frac{\overline{\Pr}(Y_{n+1} = 0 \mid X_{n+1} = 0, D)}{\overline{\Pr}(Y_{n+1} = 1 \mid
  X_{n+1} = 0, D)}  = 
\frac{p}{1-p} \cdot \frac{n^{(\vec{x},\vec{y})}_0+1}{1} \cdot 
\frac{n_1^{(\vec{x},\vec{y})} + M+1}{n_0^{(\vec{x},\vec{y})}+M+1},
$$
so that 
$$\overline{\Pr}(Y_{n+1} = 0 \mid X_{n+1} = 0, D) \mbox{\ close\ to\ } 1,$$ 
just as in Example 1(a).

Now suppose we observe $X_{n+1} = k$ for some $k \neq 0, k \neq x_i$
for $i \in \{ 1, \ldots, n\}$ (this will happen with probability
approximately $1-p$). Then we are in the situation of Example 1(b) and
we get
$$ \overline{\Pr}(Y_{n+1} = 0 \mid X_{n+1} = k, D) \approx p.$$
Note that in this case, at least under ``reasonable'' betting schemes,  
Bayes may perform better than the minimax strategy: with
probability $p$, we observe $X_{n+1} = 0$. We then predict $Y_{n+1} =
1$ with probability close to $1$, and $Y_{n+1}$ will indeed be equal
to $1$, so conditional on $X_{n+1} = 0$ we predict better than
minimax. With probability $1-p$ we observe a value of 
$X_{n+1} = k$ we typically 
have not seen before; then $Y_{n+1}$ turns out to be
$1$ and we predict $Y_{n+1} = 1$ with probability close to $p$, which
is equal to what we would have predicted if we had adopted the 
minimax strategy.

Thus, using my earlier terminology, 
if the bookie is {\em neutral}, then (for the given sample size
$n$) Bayes is ``better'' than minimax. Yet if the bookie is {\em
  semi-adversarial}, then Bayesian predictions may be problematic
here - take, for example, $p=0.7$. Then $30\%$ of the time (namely,
whenever $Y=1$) the bookie may offer a ticket which pays $1$ dollar if
$Y=0$, and costs $0.65$ cent. Then the Bayesian will be tempted to buy
such a ticket and so will loose $0.65$ dollar! - note that this will
happen $30\%$ of the time, so it's quite bad.
\commentout{ 
\paragraph{Example 1(d)}
It would be nice to have an example of Bayes suffering a lot  compared
to minimax (for example, as much as in Example (c)), but 
with a neutral bookie and bets with 
pay-offs between $0$ and $1$ (so that discrepancies between Bayes and
minimax may not be ``blown up'' by raising the stakes). I couldn't find
such an example. Maybe, such an example doesn't exist? (so Bayes
behaves better than I thought?) Not clear...
}
}

On the other hand, if 
$\cP$ is neither a singleton nor the set of all distributions with
the given marginal $\Pr_Y$, then ignoring may or may not be minimax
optimal, depending on the details of $\cP$. Even in some cases where
ignoring is not minimax optimal, it may still be a reasonable update
rule to use, because, no matter what $\cP$ is, ignoring $X$ is a {\em
  reliable\/} update rule \cite{Grunwald00a}. This means the
following:  Suppose that the loss function $L$ is known to the agent. Let
$a^*$ be the optimal action resulting from ignoring information about
$X$, that is, adopting the marginal $\Pr_Y$ as the distribution of
$Y$, independently of what $X$ was observed. Then it must be the case
that
\begin{equation}
\label{eq:reliable}
E_{\Pr_Y} [L_{a^*}] = E^{(X,Y)}_{\Pr} [L_{a^*}(X,Y)],
\end{equation}
meaning that the loss the {\em agent\/} expects to have using his
adopted action $a^*$ is guaranteed to be identical to the {\em true\/}
expected loss of the agent's action $a^*$. Thus, the quality of the
agent's predictions is exactly as good as the agent thinks they are,
and the agent  cannot be overly optimistic about his own 
performance.  Data will behave as if the agent's adopted distribution
$\Pr_Y$ is correct, even though it is not. 

This desirable property of reliability is lost when the loss function
can depend on the observation $X$. 
To understand the impact of 
this possibility, 
consider again the situation of Example~\ref{ex:nonbayes1}, except now
assume that the loss function can depend on the observation.
\xam\label{ex:nonbayes3}
As in Example~\ref{ex:nonbayes1}, assume that $\cX = \cY = \{0,1\}$,
that the agent knows that $E_{\Pr_Y}[Y] = \Pr_Y(Y = 1) = p$ for 
some fixed $p$, and 
let
$\cP$ be the set of distributions 
on $\cX \times \cY$ with marginal $\Pr_Y$.  
Now the loss function takes three arguments, where $L(i,j,k)$ is the
loss if $i$ is predicted, the true value 
of $Y$
is $j$, and 
$X = k$
is observed.
Suppose that $L(i,j,k) = (k+1)|i-j|$.  That is, if the observation is 0,
then, as before, the loss is just the difference between the predicted
value and actual value; on the other hand, if the observation is 1, then
the loss is twice the difference.  Note that, with this loss function,
it technically no longer makes sense to talk about ignoring the
information, since we cannot even talk about the optimal rule with
respect to $\Pr_Y$.  However, as we shall see, the optimal
action is still to predict the most likely value according to $\Pr_Y$.

A priori, 
there are four possible deterministic decision rules, which have the
form ``Predict $i$ if 0 is observed and $j$ if 1 is observed'', 
which we abbreviate as $\delta_{ij}$, for $i, j \in \{0,1\}$.  
It is easy to check that 
$$\begin{array}{l}
E_{\Pr}[L_{\delta_{00}}] =  \Pr(1,0) +  2 \Pr(1,1) = \Pr_Y(1) + \Pr(1,1)\\
E_{\Pr}[L_{\delta_{01}}] =  \Pr(1,0) + 2 \Pr(0,1)\\
E_{\Pr}[L_{\delta_{10}}] =  \Pr(0,0) + 2 \Pr(1,1)\\
E_{\Pr}[L_{\delta_{11}}] =  \Pr(0,0) + 2 \Pr(0,1) = \Pr_Y(0) + \Pr(0,1).
\end{array}
$$
It is not hard to show that randomization does not help in this case,
and the 
minimax
optimal decision rule is to predict 
0 if $\Pr_Y(1) = p < 1/2$ and 1 if $p > 1/2$ 
(with any way of randomizing leading to the same loss if
$p = 1/2$).  Thus, the 
minimax-%
optimal decision rule still chooses the most
likely prediction according to $\Pr_Y$, independent of the observation.

On the other hand, if either 0 or 1 is observed, the same arguments as
before show that the minimax-optimal action with respect to the conditional
probability is to randomize, predicting both 0 and 1 with probability
$1/2$.  So again, we have time inconsistency
in the sense discussed in Section~\ref{sec:nonbayes}, 
and ignoring the
information is the right thing to do.

But now consider what happens when the loss function is 
$L'(i,j,k) = (|k-j|+1)|i-j|$.  Thus, if the actual value and the
observation are the same, then the loss function is the difference
between the actual value and the prediction; however, if the actual
value and the observation are different, then the loss is twice the
difference.  Again we have the same four decision rules as above, but
now we have
$$\begin{array}{l}
E_{\Pr}[L'_{\delta_{00}}] =  2\Pr(1,0) +  \Pr(1,1) = \Pr_Y(1) + \Pr(1,0)\\
E_{\Pr}[L'_{\delta_{01}}] =  2\Pr(1,0) + 2 \Pr(0,1)\\
E_{\Pr}[L_{\delta_{10}}] =  \Pr(0,0) + \Pr(1,1)\\
E_{\Pr}[L_{\delta_{11}}] =  \Pr(0,0) + 2 \Pr(0,1) = \Pr_Y(0) + \Pr(0,1).
\end{array}
$$
Now the 
minimax-optimal rule is 
to predict 0 if $\Pr_Y(1) \le 1/3$, to predict 1
if $\Pr_Y(1) \ge 2/3$, and to use the randomized decision rule
$\frac{1}{3} \delta_{01} + \frac{2}{3} \delta_{10}$ (which has expected
loss $2/3$) if $1/3 \le \Pr_Y(1) \le 2/3$).  (In the case that $\Pr_Y(1)
= 1/3$ or $\Pr_Y(1) = 2/3$, then the two recommended rules have the same
payoff.)  

On the other hand, if $i$ is observed ($i \in \{0,1\}$), then the
minimax optimal action is to predict $i$ with probability $1/3$ and
$1-i$ with probability $2/3$. That is, the optimal strategy corresponds
to the decision rule $\frac{1}{3} \delta_{01} + \frac{2}{3}
\delta_{10}$.  Thus, in this case, there is no time inconsistency if
$1/3 \le \Pr_Y(1) \le 2/3$. 
\exam

\section{RELATED WORK}
\label{sec:related}
\commentout{
This paper is about making decisions about random variable $Y$ given
the value of $X$, when the distribution of $(X,Y)$ is in a given set
of distributions $\cP$ that all agree on a given marginal
$\Pr_Y$. We discussed three ways in which this can be done: (1) 
making minimax decisions based on the set
of conditional distributions 
 $\cP_i = \{\Pr(\cdot \mid X = i): \Pr \in
\cP\}$ as at the end of Section~\ref{sec:nonbayes}; (2) by imposing a
prior on $\cP$ and using the Bayesian predictive distribution as in
Section~\ref{sec:bayes}; and (3), by ignoring observations of $X$, effectively
restricting the sample space to $\cY$ and using $\Pr_Y$. But there are
other well-known methods to deal with such a situation, perhaps the
most popular of which is the {\em maximum entropy principle\/}
\cite{Jaynes89}:
\subsection{Ignoring and Maximum Entropy}
Suppose we are given a convex set of distributions $\cP$ on a finite
sample space $\cZ$, and we are asked to make a prediction
on a random variable $Y$ defined on $\cZ$, given the value of another
random variable $X$ on $\cZ$. The `raw' MaxEnt Principle\footnote{We
  call it `raw' to stress the difference from the Minimum Relative
  Entropy Principle, which assumes that a unique prior distribution
  $Q$ on $\cZ$ is known to the agent and $\cP$ is interpreted as {\em
    new information on which $Q$ needs to be updated}. While formally,
  raw MaxEnt is equivalent to minimum relative entropy with a uniform
  $Q$, the interpretation is quite different: the raw MaxEnt principle
  deals with situations where {\em no\/} single distribution on $\cZ$
  can be specified a priori, and $\cP$ is simply the set of
  distributions consistent with the agent's knowledge about the
  situation.}  advocates to pick the unique $P^* \in \cP$ that
maximizes, among all $P \in \cP$, the Shannon entropy $\sum_{z \in
  \cZ} - P(z) \log P(z)$, and act as if $(X,Y)$ were distributed
according to $P^*$. It can be shown, and it is well-known, that if $P
\in \cP$ is the set of distributions on $\cZ$ with a given marginal
$P_Y$, then $X$ and $Y$ are independent according to $P^*$, so that
prediction of $Y$ according to $P^*$ amounts to ignoring the value of
$X$, consistent with our recommendation in this paper.

However, if $\cP$ encodes further constraints, then it may be the case
that the MaxEnt $P^*$ does introduce a dependency
between $X$ and $Y$ after all, even though $\cP$ contains
distributions under which $X$ and $Y$ are independent. Examples are
the Pearl-Dalkey-Hunter \cite{Hunter} and the Judy Benjamin problem,
which have traditionally been interpreted as exhibiting problematic
behavior of maximum entropy. Thus, we cannot simply say that the
MaxEnt strategy tells us to ignore the value of $X$ - in some
situations it does, and in others it does not.
}
In this section we compare our work to recent and closely related work
by Seidenfeld \citeyear{Seidenfeld04}
and Augustin \citeyear{Augustin03},
as well as to various
results indicating that information should never be ignored.
\paragraph{Augustin's and Seidenfeld's work}
Seidenfeld \citeyear{Seidenfeld04} provides an
analysis of minimax decision rules which is closely related to ours,
but with markedly different conclusions. 
Suppose an agent has to
predict the value of a random variable $Y$ after observing another
random variable $X$. 
Seidenfeld  observes, as we did in Section~\ref{sec:nonbayes}, 
that the minimax paradigm can be
applied to this situation in two different ways: 
\begin{enumerate}
\item In the {\em local\/}
minimax strategy, the agent uses the minimax action relative
to the set of distributions for $Y$ conditioned on the observed value
of $X$. 
\item 
In the {\em global\/} minimax strategy, the agent adopts the
minimax decision {\em rule\/} 
(function from observations of $X$ to actions) 
relative to the set $\cP$ of joint distributions.
\end{enumerate}
Seidenfeld notes, as we did in Section~\ref{sec:nonbayes}, that
the local minimax strategy is not equivalent to the global minimax
strategy.  (In his terminology, the extensive form of the decision problem
is not equivalent to the normal form.)  Moreover, he exhibits a
rather counterintuitive property of local minimax.   Suppose that, before
observing $X$, the agent is offered the following proposition.   For an
additional small cost (loss), she will {\em not\/} be told the value
of $X$ before she has to predict $Y$. An agent who uses the local
minimax strategy would accept that proposition, because not
observing $X$ leads to a smaller minimax prediction loss than
observing $X$.
Therefore, a local minimax agent would be willing to 
pay {\em not\/} to get information.  
This is the same phenomenon that we observed in
Example~\ref{ex:nonbayes1}.

Seidenfeld interprets his observations as evidence that the
local minimax strategy is flawed, at least to some extent. 
He further views the discrepancy between local and global minimax as a
problematic aspect of the minimax paradigm. 
In a closely related context, 
Augustin \citeyear{Augustin03} also observes the discrepancy between
the global and the local minimax strategy, but, as he writes, ``there are
sound arguments for both''.

In this paper, we express a third point of view: we regard both the
strategy of ignoring information and the global minimax loss strategy
as reasonable decision rules, preferable to, for example, the local
minimax loss strategy. However, we certainly do not claim that ``global
minimax loss'' is the only reasonable strategy.  For example, as we explained
at the end of Section~\ref{sec:bayes}, in some situations minimax {\em
  regret\/} may be more appropriate. Also, as explained in
Section~\ref{sec:bookies}, if $\cP$ has a more complex structure than
the one considered in Sections~\ref{sec:nonbayes} and~\ref{sec:bayes},
then ignoring the information may no longer coincide with a global minimax
strategy.
It remains to be investigated whether, in such
cases, there is a clear preference for 
either ignoring information or for global minimax.
\paragraph{``Cost-free information should never be ignored''}
As
we observed 
in Section~\ref{sec:bayes}, a purely
subjective Bayesian who is not ``pragmatic'' in our sense
should always condition on {\em all\/} the available 
information:~information should never be ignored. 
This result
can be reconciled with our findings by noting that 
it depends
on the agent representing her uncertainty with a {\em single\/}
distribution. In the Bayesian case, the agent starts with a set of
distributions $\cP$, but this set is then transformed to a single
distribution by adopting a subjective prior on $\cP$.  The expected
value of information is then calculated using an expectation based on
the agent's prior on $\cP$. In contrast, in the ``Bayesian'' analysis
of Section~\ref{sec:bayes},
for reasons explained at the beginning of Section~\ref{sec:bayes},
we computed the expectation relative to {\em all\/} probabilities
in a set $\cP$ that is meant to represent the agent's uncertainty.
Consequently, our results differ from the subjective Bayesian analysis.
\section{Discussion}\label{sec:discussion}
We have shown that, 
in the minimax sense, 
sometimes it is better to ignore information, at
least for a while, rather than updating.
This strategy is essentially different from
other popular probability updating mechanisms such as the non-Bayesian
mechanism (local minimax) described in Section~\ref{sec:bayes} and the
Bayesian mechanism of Section~\ref{sec:nonbayes}. The
only method we are aware of that leads to similar results is the
following form of the maximum-entropy formalism: the agent first chooses
the unique distribution $P^* \in\cP$ that maximizes the Shannon entropy,
and then predicts $Y$ based on the conditional distribution $P^*(Y=
\cdot \mid X=x)$ \cite{CoverThomas}. Such an application of the Maximum Entropy
Principle will ignore the value of $X$ if $\cP$ contains {\em all\/}
distributions with the given marginal $\Pr_Y$. However, as we
indicated in Section~\ref{sec:bookies}, updating by ignoring can still
be useful if $\cP$ contains only a subset of the distributions with
given $\Pr_Y$. Yet in such cases, it is well known that 
the maximum entropy $P^*$ may introduce counterintuitive 
dependencies between $X$ and $Y$ after all, as
exemplified by the Judy Benjamin problems
\cite{GroveHalpern97}, 
thereby making the method different from merely ignoring $X$ after
all.
Our minimax-optimality results
depend on the assumption that the set of possible prior
distributions contains no information about the possible correlations
between the variable of interest and the observed variable.  In
addition, they depend on the assumption that the payoff depends only
on the actual value and the predicted value of the variable of
interest.

One way of understanding the issues involved here is in terms of
knowledge, as advocated by Halpern and Tuttle \citeyear{HT},
specifically, the knowledge of the agent and the knowledge of the
``adversary'' who is choosing the loss function.  The
knowledge of the agent is encoded by the set of possible prior
distributions.  The knowledge of the adversary
is encoded in our assumptions on the loss function.  If
the adversary does not know the observation at the time that the loss
function
is determined,
then the loss function cannot depend on the observation; if
the adversary  knows the observation, then it can.  More generally,
especially if negative losses (i.e., gains) are allowed, and 
the adversary can know the true distribution, then the adversary can
choose whether to allow the agent to play at all, depending on the
observation.  
In future work, we plan to consider the impact of allowing the
adversary this extra degree of freedom.
\subsubsection*{Acknowledgments}
We thank Teddy Seidenfeld and Bas van Fraassen for helpful discussions
on the topic of the paper.
Joseph Halpern was supported in part by NSF under grants
CTC-0208535 and ITR-0325453, by ONR under grants  N00014-00-1-03-41 and
N00014-01-10-511, by the DoD Multidisciplinary University Research
Initiative (MURI) program administered by the ONR under
grant N00014-01-1-0795, and by AFOSR under grant F49620-02-1-0101.
\bibliographystyle{chicago}
\bibliography{z,joe,bghk,refs}
\end{document}

%% file: defn.tex
\newtheorem{THEOREM}{Theorem}[section]
\newenvironment{theorem}{\begin{THEOREM} \hspace{-.85em} {\bf :} }%
                        {\end{THEOREM}}
\newtheorem{LEMMA}[THEOREM]{Lemma}
\newenvironment{lemma}{\begin{LEMMA} \hspace{-.85em} {\bf :} }%
                      {\end{LEMMA}}
\newtheorem{COROLLARY}[THEOREM]{Corollary}
\newenvironment{corollary}{\begin{COROLLARY} \hspace{-.85em} {\bf :} }%
                          {\end{COROLLARY}}
\newtheorem{PROPOSITION}[THEOREM]{Proposition}
\newenvironment{proposition}{\begin{PROPOSITION} \hspace{-.85em} {\bf :} }%
                            {\end{PROPOSITION}}
\newtheorem{DEFINITION}[THEOREM]{Definition}
\newenvironment{definition}{\begin{DEFINITION} \hspace{-.85em} {\bf :} \rm}%
                            {\end{DEFINITION}}
\newtheorem{CLAIM}[THEOREM]{Claim}
\newenvironment{claim}{\begin{CLAIM} \hspace{-.85em} {\bf :} \rm}%
                            {\end{CLAIM}}
\newtheorem{EXAMPLE}[THEOREM]{Example}
\newenvironment{example}{\begin{EXAMPLE} \hspace{-.85em} {\bf :} \rm}%
                            {\end{EXAMPLE}}
\newtheorem{REMARK}[THEOREM]{Remark}
\newenvironment{remark}{\begin{REMARK} \hspace{-.85em} {\bf :} \rm}%
                            {\end{REMARK}}
\newcommand{\thm}{\begin{theorem}}
\newcommand{\lem}{\begin{lemma}}
\newcommand{\pro}{\begin{proposition}}
\newcommand{\dfn}{\begin{definition}}
\newcommand{\rem}{\begin{remark}}
\newcommand{\xam}{\begin{example}}
\newcommand{\cor}{\begin{corollary}}
\newcommand{\prf}{\noindent{\bf Proof:} }
\newcommand{\ethm}{\end{theorem}}
\newcommand{\elem}{\end{lemma}}
\newcommand{\epro}{\end{proposition}}
\newcommand{\edfn}{\bbox\end{definition}}
\newcommand{\erem}{\bbox\end{remark}}
\newcommand{\exam}{\bbox\end{example}}
\newcommand{\ecor}{\end{corollary}}
\newcommand{\eprf}{\bbox\vspace{0.1in}}
\newcommand{\beqn}{\begin{equation}}
\newcommand{\eeqn}{\end{equation}}
\newcommand{\bbox}{\vrule height7pt width4pt depth1pt}

\newcommand{\clm}{\begin{claim}}
\newcommand{\eclm}{\end{claim}}
\newcommand{\IR}{\mbox{$I\!\!R$}}

\renewcommand{\phi}{\varphi}
\renewcommand{\P}{{\cal P}}

\newcommand{\X}{{\cal X}}
\newcommand{\Y}{{\cal Y}}

\newcommand{\ol}{\setlength{\itemsep}{0pt}\begin{enumerate}}
\newcommand{\eol}{\end{enumerate}\setlength{\itemsep}{-\parsep}}
\newcommand{\ul}{\setlength{\itemsep}{0pt}\begin{itemize}}
\newcommand{\dl}{\setlength{\itemsep}{0pt}\begin{description}}
\newcommand{\edl}{\end{description}\setlength{\itemsep}{-\parsep}}
\newcommand{\eul}{\end{itemize}\setlength{\itemsep}{-\parsep}}

\newcommand{\cA}{{\cal A}}

\newcommand{\cS}{{\cal S}}

\newcommand{\commentout}[1]{}

\newcommand{\bi}{\begin{itemize}}
\newcommand{\ei}{\end{itemize}}
\newcommand{\be}{\begin{enumerate}}
\newcommand{\ee}{\end{enumerate}}

%% file: ignore1corr.bbl
\begin{thebibliography}{}

\bibitem[\protect\citeauthoryear{Augustin}{Augustin}{2003}]{Augustin03}
Augustin, T. (2003).
\newblock On the suboptimality of the generalized {B}ayes rule and robust
  {B}ayesian procedures from the decision theoretic point of view:~{A}
  cautionary note on updating imprecise priors.
\newblock In {\em 3rd Int. Symp.~Imprecise Probabilities and Their
  Applications}, pp.\  31--45.
\newblock Available at http://www.carleton-scientific.com/isipta/2003-toc.html.

\bibitem[\protect\citeauthoryear{Barron, Rissanen, and Yu}{Barron
  et~al.}{1998}]{BarronRY98}
Barron, A.~R., J.~Rissanen, and B.~Yu (1998).
\newblock The {M}inimum {D}escription {L}ength {P}rinciple in coding and
  modeling.
\newblock {\em IEEE Trans.~Information Theory\/}~{\em 44\/}(6), 2743--2760.
\newblock Special Commemorative Issue: Information Theory: 1948-1998.

\bibitem[\protect\citeauthoryear{Bernardo and Smith}{Bernardo and
  Smith}{1994}]{BernardoS94}
Bernardo, J.~M. and A.~F.~M. Smith (1994).
\newblock {\em Bayesian Theory}.
\newblock John Wiley.

\bibitem[\protect\citeauthoryear{Blackwell and Dubins}{Blackwell and
  Dubins}{1962}]{BlackwellD62}
Blackwell, D. and L.~Dubins (1962).
\newblock Merging of opinions with increasing information.
\newblock {\em Annals of Mathematical Statistics\/}~{\em 33}, 882--886.

\bibitem[\protect\citeauthoryear{Cover and Thomas}{Cover and
  Thomas}{1991}]{CoverThomas}
Cover, T.~M. and J.~A. Thomas (1991).
\newblock {\em Elements of Information Theory}.
\newblock New York: Wiley.

\bibitem[\protect\citeauthoryear{Cozman and Walley}{Cozman and
  Walley}{2001}]{CozmanWalley}
Cozman, F.~G. and P.~Walley (2001).
\newblock Graphoid properties of epistemic irrelevance and independence.
\newblock In {\em 2nd Int.~Symp.~Imprecise Probabilities and Their
  Applications}, pp.\  112--121.
\newblock Available at http://www.sipta.org/~isipta01/proceedings/index.html.

\bibitem[\protect\citeauthoryear{Ghosal}{Ghosal}{1998}]{Ghosal98}
Ghosal, S. (1998).
\newblock A review of consistency and convergence rates of posterior
  distribution.
\newblock In {\em Proc.~Varanasi Symp.~on Bayesian Inference}.
\newblock At http://www4.stat.ncsu.edu/~sghosal/papers.html.

\bibitem[\protect\citeauthoryear{Gilboa and Schmeidler}{Gilboa and
  Schmeidler}{1989}]{GS1989}
Gilboa, I. and D.~Schmeidler (1989).
\newblock Maxmin expected utility with a non-unique prior.
\newblock {\em Journal of Mathematical Economics\/}~{\em 18}, 141--153.

\bibitem[\protect\citeauthoryear{Good}{Good}{1967}]{Good67}
Good, I. (1967).
\newblock On the principle of total evidence.
\newblock {\em The British Journal for the Philosophy of Science\/}~{\em 17},
  319--321.

\bibitem[\protect\citeauthoryear{Grove and Halpern}{Grove and
  Halpern}{1997}]{GroveHalpern97}
Grove, A.~J. and J.~Y. Halpern (1997).
\newblock Probability update: conditioning vs.~cross-entropy.
\newblock In {\em Proc.~Thirteenth Conference on Uncertainty in Artificial
  Intelligence (UAI '97)}, pp.\  208--214.

\bibitem[\protect\citeauthoryear{Gr{\"u}nwald}{Gr{\"u}nwald}{2000}]{Grunwald00%
a}
Gr{\"u}nwald, P.~D. (2000).
\newblock Maximum entropy and the glasses you are looking through.
\newblock In {\em Proc.~Sixteenth Conference on Uncertainty in Artificial
  Intelligence (UAI 2000)}, pp.\  238--246.

\bibitem[\protect\citeauthoryear{Halpern}{Halpern}{1997}]{Hal15}
Halpern, J.~Y. (1997).
\newblock On ambiguities in the interpretation of game trees.
\newblock {\em Games and Economic Behavior\/}~{\em 20}, 66--96.

\bibitem[\protect\citeauthoryear{Halpern and Tuttle}{Halpern and
  Tuttle}{1993}]{HT}
Halpern, J.~Y. and M.~R. Tuttle (1993).
\newblock Knowledge, probability, and adversaries.
\newblock {\em Journal of the ACM\/}~{\em 40\/}(4), 917--962.

\bibitem[\protect\citeauthoryear{Herron, Seidenfeld, and Wasserman}{Herron
  et~al.}{1997}]{HerronSW97}
Herron, T., T.~Seidenfeld, and L.~Wasserman (1997).
\newblock Divisive conditioning: Further results on dilation.
\newblock {\em Philosophy of Science\/}~{\em 64}, 411--444.

\bibitem[\protect\citeauthoryear{Jeffreys}{Jeffreys}{1946}]{Jeffreys46}
Jeffreys, H. (1946).
\newblock An invariant form for the prior probability in estimation problems.
\newblock {\em Proc.~Royal Statistical Society, Series A\/}~{\em 186},
  453--461.

\bibitem[\protect\citeauthoryear{Kontkanen, Myllym\"aki, Silander, Tirri, and
  Gr\"unwald}{Kontkanen et~al.}{2000}]{KontkanenMSTG00}
Kontkanen, P., P.~Myllym\"aki, T.~Silander, H.~Tirri, and P.~Gr\"unwald (2000).
\newblock On predictive distributions and {B}ayesian networks.
\newblock {\em Journal of Statistics and Computing\/}~{\em 10}, 39--54.

\bibitem[\protect\citeauthoryear{Piccione and Rubinstein}{Piccione and
  Rubinstein}{1997}]{PR97}
Piccione, M. and A.~Rubinstein (1997).
\newblock On the interpretation of decision problems with imperfect recall.
\newblock {\em Games and Economic Behavior\/}~{\em 20\/}(1), 3--24.

\bibitem[\protect\citeauthoryear{Raiffa and Shlaifer}{Raiffa and
  Shlaifer}{1961}]{RaiffaS61}
Raiffa, H. and R.~Shlaifer (1961).
\newblock {\em Applied Statistical Decision Theory}.
\newblock Cambridge, MA: Harvard University Press.

\bibitem[\protect\citeauthoryear{Seidenfeld}{Seidenfeld}{2004}]{Seidenfeld04}
Seidenfeld, T. (2004).
\newblock A contrast between two decision rules for use with (convex) sets of
  probabilities: $\gamma$-maximin versus {$E$}-admissibility.
\newblock {\em Synthese\/}.
\newblock To appear.

\bibitem[\protect\citeauthoryear{Seidenfeld and Wasserman}{Seidenfeld and
  Wasserman}{1993}]{SeidenfeldW93}
Seidenfeld, T. and L.~Wasserman (1993).
\newblock Dilation for convex sets of probabilities.
\newblock {\em Annals of Statistics\/}~{\em 21}, 1139--1154.

\end{thebibliography}
